\documentclass{bmvc2k}
\usepackage{graphicx}
\usepackage{booktabs} % For formal tables
\usepackage{bbm} % For formal tables
\usepackage{mathtools}
\usepackage{xfrac}
\usepackage{xspace}
\usepackage{wrapfig}

\usepackage{tikz}
\usepackage{comment}
\usepackage{amsmath,amssymb} % define this before the line numbering.
\usepackage{color}
\usepackage{subfig}
\newcommand{\AlgoName}{Search for Concepts: Discovering Visual Concepts Using Direct Optimization\xspace}

\DeclareMathOperator*{\argmax}{arg\,max}
\DeclareMathOperator*{\argmin}{arg\,min}

\newcommand{\preddy}[1]{\textcolor{blue}{PR: #1}}
\newcommand{\paul}[1]{\textcolor{orange}{PL: #1}}
\newcommand{\remove}[1]{\textcolor{red}{remove: #1}}

\title{\AlgoName}

\addauthor{Pradyumna Reddy}{https://preddy5.github.io/}{1}
\addauthor{Paul Guerrero}{http://paulguerrero.net/}{2}
\addauthor{Niloy J. Mitra}{http://www0.cs.ucl.ac.uk/staff/n.mitra/index.html}{1, 2}

\addinstitution{
 University College London\\
}
\addinstitution{
 Adobe Research London\\
}

\runninghead{Pradyumna Reddy, Paul Guerrero, Niloy J. Mitra}{Search for Concepts}

\definecolor{turquoise}{cmyk}{0.65,0,0.1,0.3}
\definecolor{purple}{rgb}{0.65,0,0.65}
\definecolor{dark_green}{rgb}{0, 0.5, 0}
\definecolor{orange}{rgb}{0.99, 0.39, 0.0}
\definecolor{red}{rgb}{0.8, 0.2, 0.2}
\definecolor{darkred}{rgb}{0.6, 0.1, 0.05}
\definecolor{blueish}{rgb}{0.0, 0.88, .87}
\definecolor{light_gray}{rgb}{0.7, 0.7, .7}
\definecolor{pink}{rgb}{1, 0, 1}
\definecolor{greyblue}{rgb}{0.25, 0.25, 1}

\begin{document}

\maketitle

\begin{center}
    \centering
    \vspace{-25pt}
    \includegraphics[width=\textwidth]{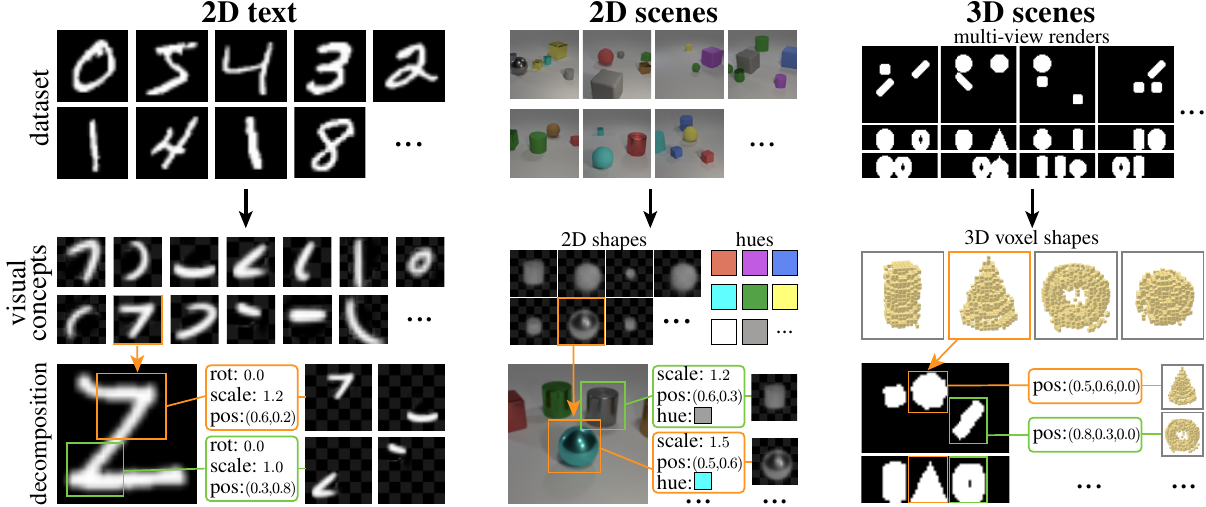} %figures/teaser/teaser.pdf}
    \vspace{0pt}
    \captionof{figure}{
    We present a search-and-learn paradigm that starts from an unlabeled dataset and a known image formation model, and learns \emph{visual concepts} in the form of a dictionary of base elements along with their placement parameters to best explain the input dataset. Here we show results on the MNIST dataset, Clevr renderings, and a 3D sprite dataset. 
    }
    \label{fig:teaser}
    %\vspace{-5pt}
\end{center}%

\begin{abstract}
Finding an unsupervised decomposition of an image into individual objects is a key step to leverage compositionality and to perform symbolic reasoning.
Traditionally, this problem is solved using amortized inference, which does not generalize beyond the scope of the training data, may sometimes miss correct decompositions, and requires large amounts of training data.
%\remove{Traditionally, this problem is solved as supervised a  segmentation task, which does not generalize beyond the scope of the supervision. Hence, unsupervised methods have rercently been proposed that train a network to decompose a scene into a sparse set of objects, but their entangled latent representation hides information about the objects.}
%
We propose finding a decomposition using direct, unamortized optimization, via a combination of a gradient-based optimization for differentiable object properties and global search for non-differentiable properties.
%using sets of parameterized objects that enables unsupervised decomposition with a direct optimization over parameter space \remove{instead of employing a network.} \add{ and we investigate the limits of such a model.}
%
We show that using direct optimization is more generalizable, misses fewer correct decompositions, and typically requires less data than methods based on amortized inference.
This highlights a weakness of the current prevalent practice of using amortized inference that can potentially be improved by integrating more direct optimization elements. 
%this both improves the decomposition performance significantly, and generates additional information about the objects, in the form of visual concepts, that is not available to the network-based methods, highlighting a
%weakness of the current prevalent practice that can potentially be improved by integrating more direct optimization
%elements. 
%In this work we tackle the difficult problem of given a composition function and input signal, in an unsupervised manner discovering the parts and their parameters to reconstruct input signal using the composition function.
\end{abstract}
%\vspace{-10pt}
\section{Introduction}
%\vspace{-5pt}
% long standing problem
Reconstructing an input signal as a \textit{composition} of different meaningful parts is a long standing goal in data analysis. The ability to decompose a signal into meaningful parts not only results in an interpretable abstraction, but also improves sampling efficiency and generalization of learning-based algorithms. 
Notable classical unsupervised methods for part/parameter decomposition include Principal Component Analysis (PCA), Independent Component Analysis (ICA), Dictionary Learning, Matching pursuits.
%(see Figure~\ref{fig:pattern_oldschool_decomp} for a comparison).
In computer vision, the output of these methods are regularly used for classification, denoising, texture propagation, etc.

In the context of images, amortised optimization with neural networks is currently the unquestioned practice in self-supervised decomposition~\cite{burgess2019monet,greff2019multi,locatello2020object,smirnov2021marionette}. Amortised optimization is fast and has the potential to avoid local minima, but can be inexact and is known to struggle with more complex settings. For this reason, several well-known works like AlphaZero, AlphaGo, and AlphaFold mix \emph{direct search} with amortised optimization.

We pose the question \textit{if direct optimization can also benefit unsupervised scene decomposition}. In this paper, we learn unsupervised \textit{visual concepts} from data using a direct search approach instead of amortized inference.
By visual concepts, we refer to a small dictionary of (unknown) parameterized objects, that are acted upon by parameterized transformations (e.g., translation, rotation, hue change), resulting in transformed instances of the visual concepts called \emph{elements}, which are rendered into a final image using a given image formation model.
Given access to a sufficiently large dataset, we demonstrate that interpretable visual concepts naturally emerge as they allow efficient explanation of diverse datasets. 

While the direct search problem for visual concepts is computationally ill-behaved, we show that splitting the problem into subtasks not only results in computationally efficient problems but also provides, as empirically observed, near optimal solutions. Particularly, we alternate between solving for the dictionary of visual concepts and their parameterized placement across any given image collection. 
We show that this approach has several advantages: using an optimization to perform the decomposition, instead of a single forward pass in a network, (i)~allows finding solutions that the network missed and improves decomposition performance and (ii)~often requires less training data than amortized inference and produces (iii)~fully interpretable decompositions where elements can be edited by the user.
%The resultant parametric elements have several advantages: (i)~our parametric elements allow directly optimizing for the element parameters, unlike traditional methods that use a forward pass in a trained network to perform the decomposition; (ii)~using an optimization to perform the decomposition, instead of a single forward pass in a network, allows finding solutions that the network missed and improves decomposition performance; and finally, (iii)~produce explicit interpretable information about the discovered visual concepts and the parameters of each element, which are not available in traditional network-based approaches.
For example, in Figure~\ref{fig:teaser}, our method extracts strokes from MNIST digits, 2D objects from Clevr images, and 3D objects from a multi-view dataset of 3D scene renders.
%While the first two examples are in 2D (image domain), the third example is in 3D.
%In Figure~\ref{fig:pattern_oldschool_decomp}, we compare to classical unsupervised methods.

%We propose an unsupervised method to decompose a signal into a set of constituent parts, giving an explicit representation of both the parts and their composition parameters.
%given a composition function allows you to learn the parts that form the input signal and their composition parameters.
We evaluate on multiple data modalities, report favorable results against different SOTA methods on multiple existing datasets, and extract interpretable elements on datasets without texture cues where deep learning methods like Slot Attention suffer. Additionally, we show that our method improves generalization performance over supervised methods. 
\section{Related Work}
%\vspace{-5pt}
%We roughly divide the methods that perform decomposition into supervised methods and unsupervised methods. 

\paragraph{Supervised methods.}
The well-studied problems of instance detection and semantic segmentation are common examples of supervised decomposition approaches. Due to the large body of literature, we only discuss some representative examples. 
%Different Supervised methods have posed the problem of decomposition differently.
% 
Methods like Segnet~\cite{badrinarayanan2015segnet} and many others~\cite{chen2017deeplab,zhao2017pyramid} have tackled the problem of semantic segmentation, decomposing an image into a set of non-overlapping masks, each labelled with a semantic category.
%classifying each pixel into one of the pre-defined semantic classes.
%
Instance detection methods~\cite{girshick2014rich,girshick2015fast,ren2015faster,redmon2015you} decompose an image into a set of bounding boxes, where each box contains a semantic object, while others~\cite{krishna2016visual,lu2016visual,chen2019scene} go further by detecting relationship edges between objects, producing an entire scene graph.
%have trained networks to predict the location of each object instance in the input image using bounding boxes. 
%
%Methods like \cite{krishna2016visual, lu2016visual, chen2019scene} have proposed approaches to estimate a scene graph of an input image. Such methods require large amounts of manually annotated data and training such networks using limited data is still an active field of research.
Mask-RCNN~\cite{he2017mask} proposed an architecture to perform both instance detection and the segmentation using a single network. 
Recently, there has been a rise of architectures that that explore set generation methods~\cite{kosiorek2020conditional,zhang2019deep,lee2019set,carion2020end} for decomposition.
%DETR \cite{carion2020end} decomposes a scene into objects using a transformer.
%by estimating the set of objects present in the input image.
However, as the name indicates, supervised methods require access to different volumes of annotated data for supervision, and often fail to generalize to unseen data, beyond the scope of the distribution available for supervision.
%compared to unsupervised methods. 
%\vspace{-10pt}
\paragraph{Unsupervised methods.}
Prior to the rise of deep learning, methods~\cite{jojic2003epitomic} have been proposed to model an input signal as a composition of epitomes, which contain information about shape and appearance of objects in an input image, and further research also tried to represent objects and scenes as hierarchical graphs composed of primitives and their relationships~\cite{chen2007rapid,zhu2008unsupervised,zhu2010part,zhu2007stochastic}.
%In \cite{zhu2007stochastic} the authors present a more in-depth review and a perspective of the idea of visual grammar to represent an image using a context sensitive representation. => I didn't find anything about a grammar in the paper

%More recently, in the context of program synthesis, researchers~\cite{lake2015human,ellis2019write,ellis2020dreamcoder} have tried to decompose written letters or drawings into the primitives and operations of a domain-specific language (DSL).
%identify a set of high-order programming primitives and their usage using a Bayesian Program Learning framework to explain written letters or drawings.
%While such approaches are able to show impressive performance in program inference, scaling these methods to more general settings of large image datasets, with significant variance, is non-trivial.
%due to their inherent complexity and the specialization of the DSL to specific domains.
%For example, in the case of hand-written text, the dictionary of atomic strokes is not predefined and has to be discovered from the input. 

Several methods~\cite{sabour2017dynamic,kosiorek2019stacked,hinton2018matrix,goyal2019recurrent,locatello2020object,engelcke2021genesis} try to perform decomposition as routing in an embedding space.
%Each method was able to perform decomposition with varying amount of success.
The decomposition performance of these methods is sensitive to the
%hit or miss and is heavily dependant on
input data distribution and may completely fail on some common cases, as we show in Section~\ref{sec:results}.
Recently, a method
%for using \cite{locatello2020object} 
to decompose a 3D scene into multiple 3D objects was proposed~\cite{stelzner2021decomposing}. However, the method is domain specific to 3D data.
In a related line of research~\cite{greff2019multi,burgess2019monet,greff2017neural,engelcke2019genesis}, methods naturally encourage decomposing the input image into desirable sets of objects during learning. However, these methods are currently out-performed in most tasks by embeddings-space routing methods such as Slot Attention~\cite{locatello2020object}, and extending these methods to other domains is not straight forward. 
A differentiable decomposition method was recently proposed~\cite{reddy2020discovering}, however, extensive information about the content of the decomposed elements is needed as input.
%present a method which can discover the composition parameters using gradient descent given a crop the individual parts that form the image, however such a pipeline cannot be scaled easily.
%Invoking spatio-temporal reasoning, a recent method uses motion as queue to learn image decompositions~\cite{sabour2021unsupervised}, requiring datasets with moving objects. 

Inspired by the use of compositionality in traditional computer graphics pipelines, recent generative methods for 3D scenes encourage object-centric representations, using 3D priors~\cite{nguyen2020blockgan,nguyen2019hologan,Niemeyer2020GIRAFFE,EhrhardtGrothEtAl:RELATE:NeurIPS:2020,van2020investigating}. However, such ideas are yet to be extended beyond the generative setting.
Decomposition is also discussed in more general AI-focused contexts~\cite{greff2020binding}.
Most recently, DTI-Sprites\cite{monnier2021unsupervised}, Marionette~\cite{smirnov2021marionette} use a neural network to estimate a decomposition into a set of learned sprites, however the reliance on differential sampling and soft occlusion introduces local minima and undesirable artifacts.  
%the performance of this methods in the absence of color queues is yet to be demonstrated. 

%\paul{Need to add Marionette~\cite{smirnov2021marionette} and InSeGAN~\cite{cherian2021insegan}.}

%\remove{We also encourage compositionality of the elements, by explicitly fixing the choice of the compositionality operator (e.g., alpha composite, 3D projection) and demonstrate that, with this simple choice and by encouraging re-use, it is possible to discover compact dictionaries of interpretable atomic visual concepts in various domains (e.g., hand-written letters, 2D scenes, 3D scene renderings).}

%\vspace{-10pt}
\section{Overview}
%\vspace{-5pt}
% \input{includes/tables/notations_table}

\begin{figure*}[t!]
\centering
  \includegraphics[width=\textwidth]{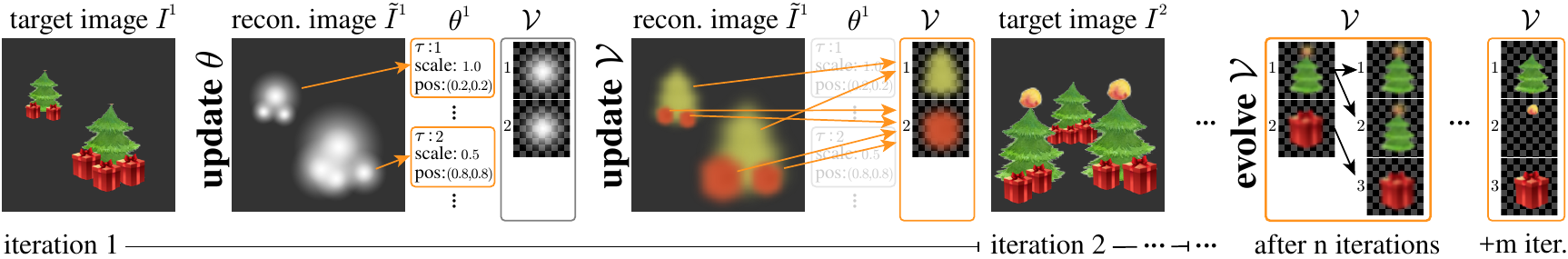}%
  \vspace{4pt}
  \caption{Overview of our approach. Given an image dataset $\mathcal{I} = \{I^1, \dots, I^N\}$, we alternate between updating element parameters $\theta$ (such the choice of visual concept $\tau$, its position, scale, etc.) and the visual concepts $\mathcal{V}$. We iterate over multiple target images $I^k$ before evolving $\mathcal{V}$ by cloning concepts that are used often and have a large error. Another $m$ iterations specialize the cloned concepts to better reconstruct the target images. The result of this iterative procedure is a library of visual concepts that can be used to efficiently decompose any image that makes use of similar visual concepts.}
  \label{fig:pipeline}
  \vspace{-15pt}
\end{figure*}

% Our goal is unsupervised decomposition of an RGBA image $I$ into a set of elements $L_1, \dots, L_n$, that approximate $I$ when combined using alpha compositing and where each element contains a single \emph{visual concept}.
Our goal is an unsupervised decomposition of an RGBA image $I$ into a set of elements $E_1, \dots, E_n$ that approximate $I$ when combined using a given image formation function and where each element is an instance of a \emph{visual concept}.
A visual concept is an  (unknown) object or pattern that commonly occurs in a dataset of images $\mathcal{I}$, such as Tetris blocks in a dataset of Tetris scenes, characters in a dataset of text images, or individual strokes in a dataset of hand-drawn characters.
Figure~\ref{fig:pipeline} shows an overview of our approach. 
%semantically meaningful object or image part, and that.
%For a single image, this task is generally ill-defined, as it is unclear which parts of the image correspond to semantica
%We define semantically meaningful objects in a data-driven context, as objects commonly occurring in a large dataset of images $\mathcal{I}$.

An \emph{element} $E_i$ is a transformed instance of a visual concept. We use a parametric function $e$ to create each element $E_i = e_\mathcal{V}(\theta_i)$, where $\mathcal{V}$ is a sparse dictionary of \textit{visual concepts} extracted from an image dataset $\mathcal{I}$, and $\theta_i = (\tau_i, \phi_i)$ is a set of \textit{per-element parameters}: $\tau_i$ is an index that describes which visual concept from $\mathcal{V}$ element $E_i$ is an instance of, and $\phi_i$ are domain-specific transformation parameters, such as translations, rotations and scaling. The reconstructed image $\tilde{I} = h(E_1, \dots , E_n)$ is computed from the elements with an image formation function $h$, which we assume to be given and fixed. Details on the parametric image and element representations are given in Section~\ref{sec:elements}. 

We learn visual concepts $\mathcal{V}$ by optimizing both $\mathcal{V}$ and the element parameters $\theta_i$ to reconstruct an image dataset $\mathcal{I}$. The dictionary $\mathcal{V}$ is shared between all images in $\mathcal{I}$, while $\theta_i$ has different values for each element. While jointly optimizing $\mathcal{V}$ and $\theta_i$ jointly is hard, we find that that optimizing one given the other is tractable. Thus, we alternate between optimizing $\mathcal{V}$ and $\theta_i$. At test time, we keep the visual concepts fixed and only optimize for the element parameters that best reconstruct a given image. The optimization is described in Section~\ref{sec:optimization}.

In Sections~\ref{sec:elements} and~\ref{sec:optimization}, we first describe our approach with 2D alpha compositing as image formation function, and 3D voxel compositing is described in the supplement.
\section{Parametric Elements and Images Formation}
%\vspace{-5pt}
\label{sec:elements}
We approximate an image $I$ with a set of parametric elements $\tilde{I} = h(E_1, \dots, E_n)$, where each element is an instance of a visual concept.

%\vspace{-10pt}
\paragraph{Visual concepts.}
The dictionary of visual concepts $\mathcal{V} = (V_1, \dots, V_m)$ defines a list of visual building blocks that can be transformed and arranged to reconstruct each image in a dataset $\mathcal{I}$. A visual concept $V_j$ is defined as a small RGBA image patch of a user-specified size
%\preddy{each visual concept can be future parameterized, for example into textures and shapes}.
The size of the patch determines the maximum size of a visual concept. Depending on the application, we either set the number $m$ of visual concepts manually, or learn the number while optimizing the dictionary. Section~\ref{sec:optimization} provides more details on the optimization.

%\vspace{-10pt}
\paragraph{Parametric elements.}
Each element $E_i = e_\mathcal{V}(\theta_i)$ is a transformed visual concept. The parameters $\theta_i = (\tau_i, \phi_i)$ determine which visual concept is used with $\tau_i \in [1, m]$, and how the visual concept is transformed with the parameters $\phi_i$:
%\begin{equation}
    $E_i = e_\mathcal{V}(\theta_i) \coloneqq T(V_{\tau_i}, \phi_i),$
%\end{equation}
where $T(V, \phi_i)$ transforms a visual concept $V$ according to the parameters $\phi_i$ and re-samples it on the image pixel grid (samples that fall outside the area of the visual concept have zero alpha and do not contribute to the final image). The type of transformations performed depend on the application, and may include translations, rotations and scaling. See Section~\ref{sec:optimization} for details. % on specific applications.

%\vspace{-10pt}
\paragraph{Image formation function.}
The reconstructed image $\tilde{I}$ is an alpha-composite of the individual elements:
%\vspace{-8pt}
\begin{equation}
    \tilde{I} = h(E_1, \dots, E_n) \coloneqq \sum_{i=1}^n E_i \prod_{j=1}^{i-1} ( 1 - E_j^{\text{A}} )
    \label{eq:alpha_compositing}
\end{equation}
where
%$\text{X} \in \{\text{R}, \text{G}, \text{B}, \text{A}\}$ denotes a single channel and
$\text{A}$ is the alpha channel and channel products are element-wise (with broadcasting to avoid a cluttered notation). We set the maximum number of elements $n$ manually ($n$ is between 4 and 45 in our experiments, depending on the dataset). Note that we can also use fewer elements than the maximum since the transformation $T$ can place elements outside the image canvas, where they do not contribute to the image.

%\vspace{-10pt}
\section{Optimizing Visual Concepts}
%\vspace{-5pt}
\label{sec:optimization}

We train our dictionary of visual concepts to reconstruct a large image dataset $\mathcal{I}$ as accurately as possible:
%\vspace{-16pt}
\begin{equation}
    \argmin_{\mathcal{V}, \Theta} \mathcal{E}(\mathcal{V}, \Theta) \coloneqq \sum_{k=1}^{|\mathcal{I}|} \|I^k - \tilde{I}^k\|_2^2  %\\ \nonumber
    \quad  \text{with } \tilde{I}^k = h(e_\mathcal{V}(\theta_1^k), \dots, e_\mathcal{V}(\theta_n^k)),
\end{equation}
where $\mathcal{E}$ is the $L_2$ reconstruction error, $\theta_i^k$ denotes the parameters of element $i$ in image $k$, and ${\Theta} = \{ \theta_i^k \}$ is the set of element parameters in all the images of $\mathcal{I}$. Optimizing over $\mathcal{V}$ and ${\Theta}$ jointly is infeasible since the search space is not well-behaved. It is high-dimensional, and contains both local minima and discrete dimensions, such as those corresponding to the visual concept selection parameters $\tau_i$. However, since the element parameters $\theta^k_i$ of different images $k$ appear in separate linear terms, they can be independently optimized \textit{given} $\mathcal{V}$. This motivates a search strategy that iterates over images $I_k$,  and alternates between updating the visual concepts $\mathcal{V}$ and the element parameters $\theta^k_i$.

%\vspace{-10pt}
\subsection{Updating Element Parameters}
\begin{figure}[t]
% \begin{wrapfigure}[13]{rh}{0.6\textwidth}
\centering
    % \vspace*{-.4in}
  \includegraphics[width=\linewidth]{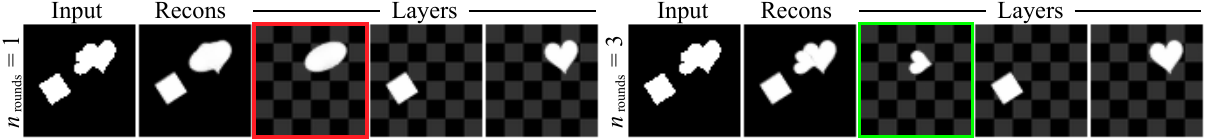}%
  \vspace{4pt}
  \caption{\label{fig:refinementWandWO} 
  Optimized element parameters after $n_\text{rounds}$ optimization rounds.\iffalse{After one round, the first layer greedily approximates both heart shapes with a single ellipse. Such greedy errors are resolved after additional rounds.}\fi
  %Showing the progress of the optimization over different rounds. Note that at the end of the first round, the reconstruction of base elements are not ideal, resulting in reconstruction error. However, as rounds progress %(i.e., the dictionary has chance to evolve over multiple images),
  %the reconstruction improves. Note that since we do not handle scaling in this example, the hearts are identified as different elements across scales. 
  }
% \end{wrapfigure}
\vspace{-15pt}
\end{figure}

%\vspace{-5pt}
While the element parameters of different images can be optimized independently, the optimal parameters of different elements in the same image depend on each other due to the alpha-composite. One possible approach to optimize all element parameters in an image given the visual concepts $\mathcal{V}$ is to use differentiable compositing~\cite{reddy2020discovering}. However, we show that even a simpler greedy approach gives us good results. We initialize all elements to be empty and optimize the parameters of one element at a time, starting at $\theta_1$. The optimum of $\theta_1$ is likely to be the least dependent on the other elements, since it corresponds to the top-most element that is not occluded by other elements. We perform $n_{\text{rounds}}$ rounds of this per-element optimization (typically $n_{\text{rounds}}=3$ in our experiments). In Figure~\ref{fig:refinementWandWO}, we compare $n_\text{rounds}=1$ versus $n_\text{rounds}=3$.
%, where the second round can take the previous results of all elements into account. 

The parameters in a single element determine the choice of visual concept in an element and its transformation. Gradient descent is not well suited for finding the element parameters, due to discontinuous parameters and local minima, but the dimensionality of the parameters is relatively small, between $2$ and $4$ in our applications. This allows us to perform a grid search in parameter space (see the supplementary material for grid resolutions). From the element parameters we typically use, the objective values are most sensitive to the translation parameters. A small translation can misalign a visual concept with the target image and cause a large change in the objective, requiring us to use a relatively high grid resolution. Fortunately, we can speed up the search over the translation parameters considerably by approximating the original objective with a normalized correlation and formulating the grid search over translations as a convolution, which can be performed efficiently with existing libraries. Details on our convolution-based grid search are given in the supplement.

% The translation of a visual concept in a parametric element is defined as $T^{\text{translation}}(V,t_x,t_y) = V($

\paragraph{Element shuffling.}
%\vspace{-10pt}
%\paul{TODO: describe additional strategies: element sorting, conditional redo}
We shuffle the order of elements to improve convergence  and avoid local minima  encountered due to our greedy per-element optimization. After optimizing all elements in an image, we move each element to the front position in turn, effectively changing the occlusion order.
% We keep each swap only if it improves the objective.
After each swap, we check the objective score and keep the swap only if it improves the objective.

\subsection{Updating Visual Concepts}
%\vspace{-5pt}
The dictionary of visual concepts $\mathcal{V}$ is shared across all images in the dataset $\mathcal{I}$. A parametric element $E_i = T(V_{\tau_i}, \phi_i)$ is differentiable w.r.t. the visual concept $V_{\tau_i} \in \mathcal{V}$, thus we can update $\mathcal{V}$ using stochastic gradient descent. After updating all element parameters of a given image $I \in \mathcal{I}$, we jointly update the visual concepts used in all elements of the image by taking one gradient descent step with the following objective:
%\begin{equation}
    $\argmin_\mathcal{V} \|I - h(e_\mathcal{V}(\theta_1), \dots, e_\mathcal{V}(\theta_n))\|^2_2,$
%\end{equation}
while keeping the element parameters $\theta_i$ fixed. To restrict the value domain of visual concepts $V_j$ to the range $[0,1]$, we avoid functions that have vanishing gradients and use an approach 
%we want to avoid functions that have vanishing gradients in large parts of their domain, like the Sigmoid.
inspired by periodic activation functions~\cite{sitzmann2019siren}: $V_j = \sin(30 V'_j)*0.5 + 0.5$, and we optimize over $V'_j$.
% \preddy{TODO: need to metion that visual concepts are projected into 0-1 range using a $sin(30*\mathcal{V})*0.5 + 0.5$}
%We use Stochastic Gradient Descent (SGD) to optimize the dictionary $\mathcal{V}$ one image batch at a time.

% \input{includes/images/cloningWandWO}

\paragraph{Evolving visual concepts.}
%\vspace{-10pt}
In many practical applications, we may not know the optimal number of visual concepts in advance. Choosing too many concepts may result in less semantically meaningful concepts, while too few concepts prevent us from reconstructing all images. We can learn the number of concepts along with the concepts using an evolution-inspired strategy. We start with a small number of visual concepts, and every $n_{\text{ev}}$ epochs, we check how well each concept performs ($n_\text{ev}$ is between $1$ and $3$ in our experiments, depending on dataset size). We replace concepts that incur a large reconstruction error and occur frequently with two identical child concepts. In the next epoch, these twin concepts will be used in different contexts, and will specialize to different patterns or objects in the images. Concepts that occur too infrequently, are removed from our dictionary. This results in a tree of visual concepts that is grown during optimization. The supplement describes thresholds for removing and splitting concepts and an illustration of the visual concept tree.

\paragraph{Composite visual concepts.}
%\vspace{-12pt}
A visual concept may appear in several discrete variations in the image dataset. For example, each Tetris block in the Tetris dataset may appear in one of 6 different hues. To avoid having to represent each combination of hue and block shape as separate visual concept, we could add a hue parameter to our element parameters. However, that would not give us explicit information about the discrete set of hues that appear in the dataset. Instead, we can split our library of visual concepts into two parts: $\mathcal{V}^s$ captures the discrete set of shapes in the dataset, and $\mathcal{V}^h$ captures the discrete set of hues as a dictionary of 3-tuples. The visual concept selector $\tau = (\tau^s, \tau^h)$ in the element parameters is then a tuple of indices, one index into the shapes and one into the hues, and the transformation function $T(V^s_{\tau^s}, V^h_{\tau^h}, \phi)$ combines shape $V^s \in \mathcal{V}^s$ with hue $V^h \in \mathcal{V}^h$ through multiplication. When using these composite visual concepts, both shape and hue dictionaries are updated in the visual concept update step.

\section{Results and Discussion}
%\vspace{-5pt}
We demonstrate our method's performance on three tasks: (i) \emph{unsupervised object segmentation}, where our unsupervised decomposition is used to segment an image that has a known ground truth segmentation, (ii) \emph{cross-dataset reconstruction}, where we test the generalization performance of our method by training our visual concepts on one dataset and using them to decompose an image from a second dataset, and (iii) \emph{3D scene reconstruction}, where we learn 3D concepts and reconstruct 3D scenes from multiple 2D views.

\if0
\paragraph{Optimization details.}
%\vspace{-10pt}
%In all our experiments
We learn visual concepts using AdaDelta~\cite{zeiler2012adadelta} with a learning of $1.0$ and a batch size of $8$ on a single GPU. 
We have tested different optimizers but found that our pipeline was robust to the choice of optimizer. A more detailed comparison of optimizers is  in the supplemental.
%a graph of mse and the choice of optimizers in the supplementary.
We do not use any learning rate schedulers or warm-up techniques.
\fi
\label{sec:results}

%We also show the ARI value of the baselines and our method when we add various degrees of synthetic noise to the datasets.
%\preddy{we need mention that we dont work with multi-dsprites since colors are randomly sampled and the search space is too big}
%\vspace{-10pt}
\subsection{Object Segmentation}
%\vspace{-5pt}
In this experiment, we measure the quality of our decompositions by comparing the segmentation induced by a decomposition to a known ground truth.

%\vspace{-10pt}
\paragraph{Datasets} \label{datasets}
%\vspace{-5pt}

We test on four decomposition datasets: Tetrominoes,  Multi-dSprites~\cite{dsprites17}, Multi-dSprites  adversarial, and Clevr6~\cite{johnson2017clevr}. Please refer to \ref{Sec:Hyperparameters} for more details on datasets, optimization parameters and other hyperparameters.
\iffalse
The Tetrominoes dataset contains $60$k images with 3 randomly rotated and positioned Tetris blocks. 
The Multi-dSprites dataset consists of $60$k images, each showing 2-3 shapes picked from a dictionary and placed at random locations, possibly with occlusions. We use a discretized variant of the color version and the binarized version of this dataset. We discretize the color version to eight colors, so that the colors can be discovered as composite concepts. In addition, we create a variant of this dataset that we name \emph{Multi-dSprites adversarial}, where shapes can only appear in three discrete locations on the canvas. This intuitively simple dataset highlights shortcomings in existing methods.
% We show that this dataset is difficult to handle for some existing methods.
The Clevr6 dataset consists of $35$k rendered 3D scenes with each scene is an arrangement of up to six geometric primitives with various colors and materials. Also since MarioNette data is not available, we obtain comparable data through screen captures of \href{https://www.youtube.com/watch?v=aj43i9Az4PY}{Space Invaders}, \href{https://www.youtube.com/watch?v=rLl9XBg7wSs}{Super Mario Bros}, and \href{https://archive.org/details/irobotnovel/page/n7/mode/2up}{I, Robot} (in the latter we use only lower-case letters). 
\fi

%\vspace{-10pt}
\paragraph{Baselines}
%\vspace{-5pt}
We compare our results with the state-of-the-art in unsupervised decomposition: Iodine~\cite{greff2019multi}, Slot Attention~\cite{locatello2020object}, DTI-Sprites~\cite{monnier2021unsupervised} and Marionette~\cite{smirnov2021marionette}, which use trained neural networks to perform the decomposition. Note that these baselines do \emph{not} create an explicit dictionary of visual concepts except Marionette. While DTI-Sprites does create a dictionary, but individual slots entangle multiple different concepts, and the trained network is needed to disentangle them at inference time. Further, Iodine and Slot Attention do not generate explicit element parameters. 

%Since objects in Multi-dSprites color are create by sampling random colors this creates an extremely large search space to optimize the decomposition making it computationally infeasible for our pipeline. 

\begin{figure}[t]
  \centering
  \subfloat
  %[Example decomposition on the Tetris dataset, showing the learned dictionary.
  %Since we do not account for reflections, mirrored elements appear as separate dictionary elements.]
  {\includegraphics[width=5.cm]{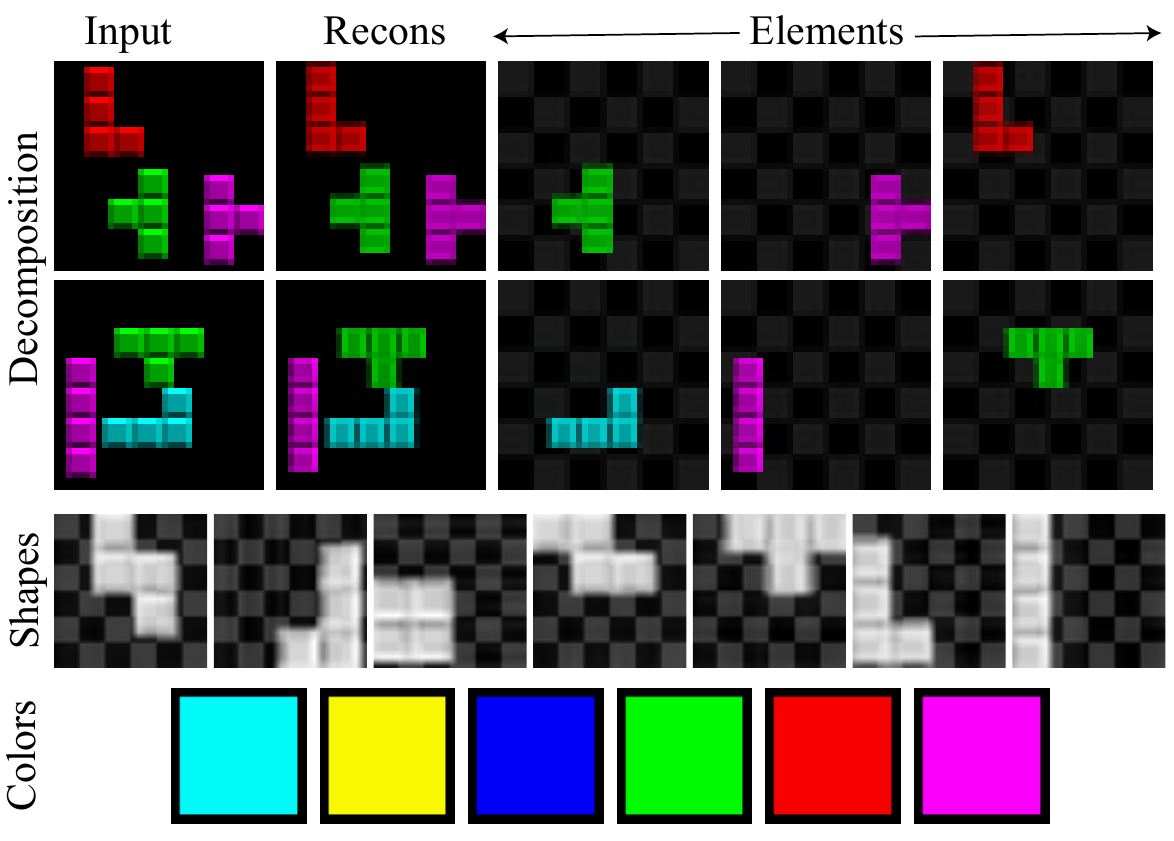} \label{fig:tetris_results}}
  \qquad
  \subfloat
  %results showing both the extracted dictionary (shapes) as well as their parameters (colors). Note that, since this analysis is in the image space, our algorithm picks up identical (3D) objects at different depths as different (image space) elements. Shadows are modelled using fuzzy element boundaries.]
  {\includegraphics[width=5.cm]{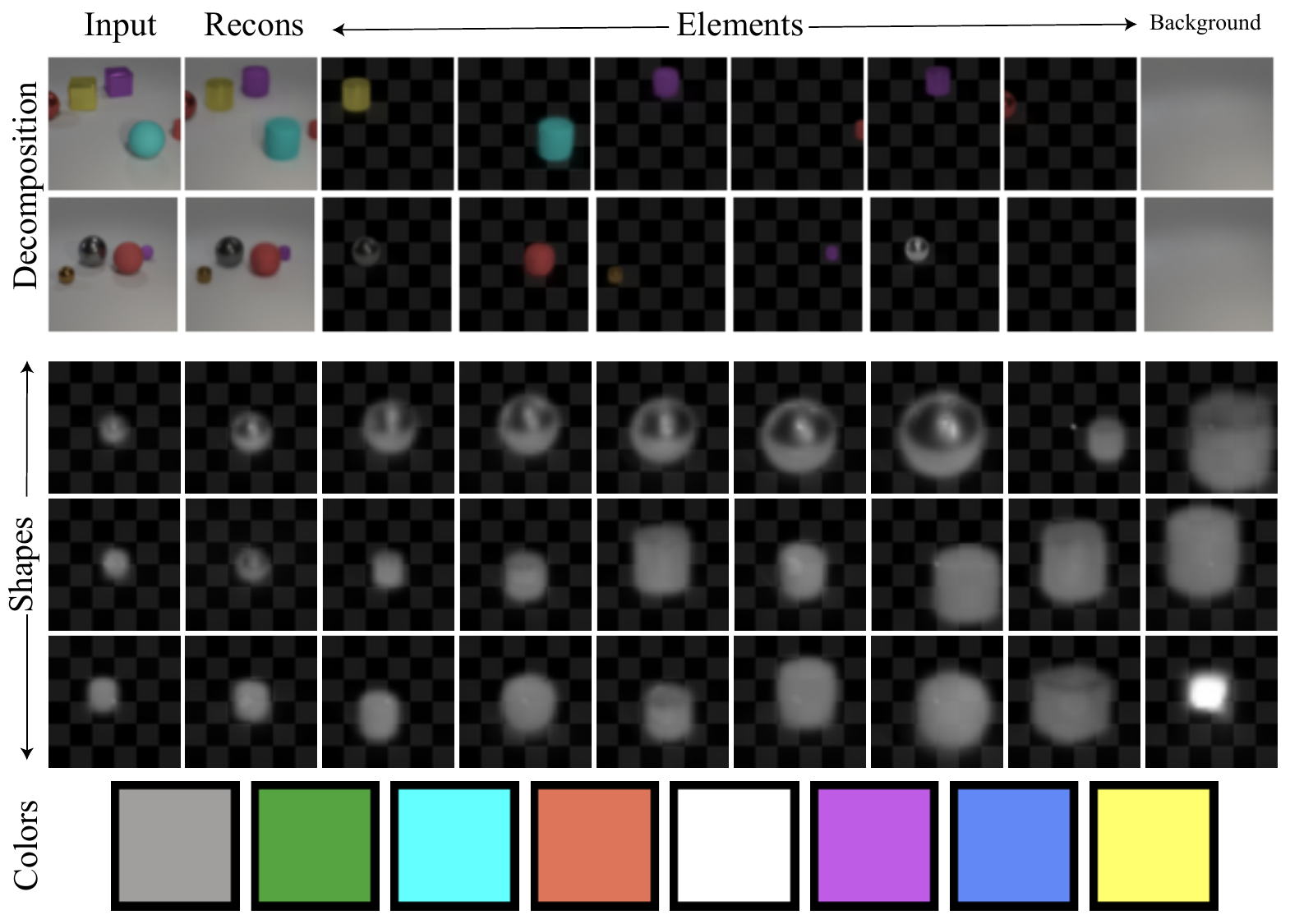} \label{fig:clevr_results}}
\vspace{3pt}
\caption{Decomposition result on the Tetris dataset (left) and Clevr dataset (right), showing composite concepts consisting of both learned shapes and learned hues. Since we do not include mirroring or scaling in our image formation model, mirrored objects and objects at different depths are learned as separate concepts.}
\label{fig:tetris_and_clevr_results}
\vspace{-15pt}
\end{figure}

% \input{includes/images/tetris_results}
%\input{includes/images/dsprites_bin_results}
%\input{includes/images/dsprites_custom_results}

%\vspace{-10pt}
\paragraph{Results}
%\vspace{-5pt}
For all the datasets, we start with $3$ visual concepts and evolve more concepts as needed. Table~\ref{tbl:ari} shows quantitative comparisons, measuring the segementation performance with the Adjusted Rand Index (ARI)~\cite{vinh2010information}. Our method achieves the best performance in the two variants of the M-dSprites dataset, with SlotAttention failing on the adversarial version. In Tetrominoes, the performance of all methods is near optimal. In the Clevr6 dataset, the lighting, reflections, and perspective projection effects violate our assumptions about the image formation model, which we assume to be alpha-blending. Nevertheless, we include Clevr6 to show that our method gracefully fails if our assumptions about the image formation do not hold.
Table~\ref{tbl:comparison} shows that the visual concepts learned by our method are closer to the ground truth concepts than for existing methods, that is, our method finds the dictionary of objects that scenes are composed of more accurately.
%
%\begin{figure}[b!]
%\begin{wrapfigure}[16]{0.6\textwidth}
\begin{wrapfigure}{r}{0.6\textwidth}
\vspace{-15pt}
\centering
  \includegraphics[width=\linewidth]{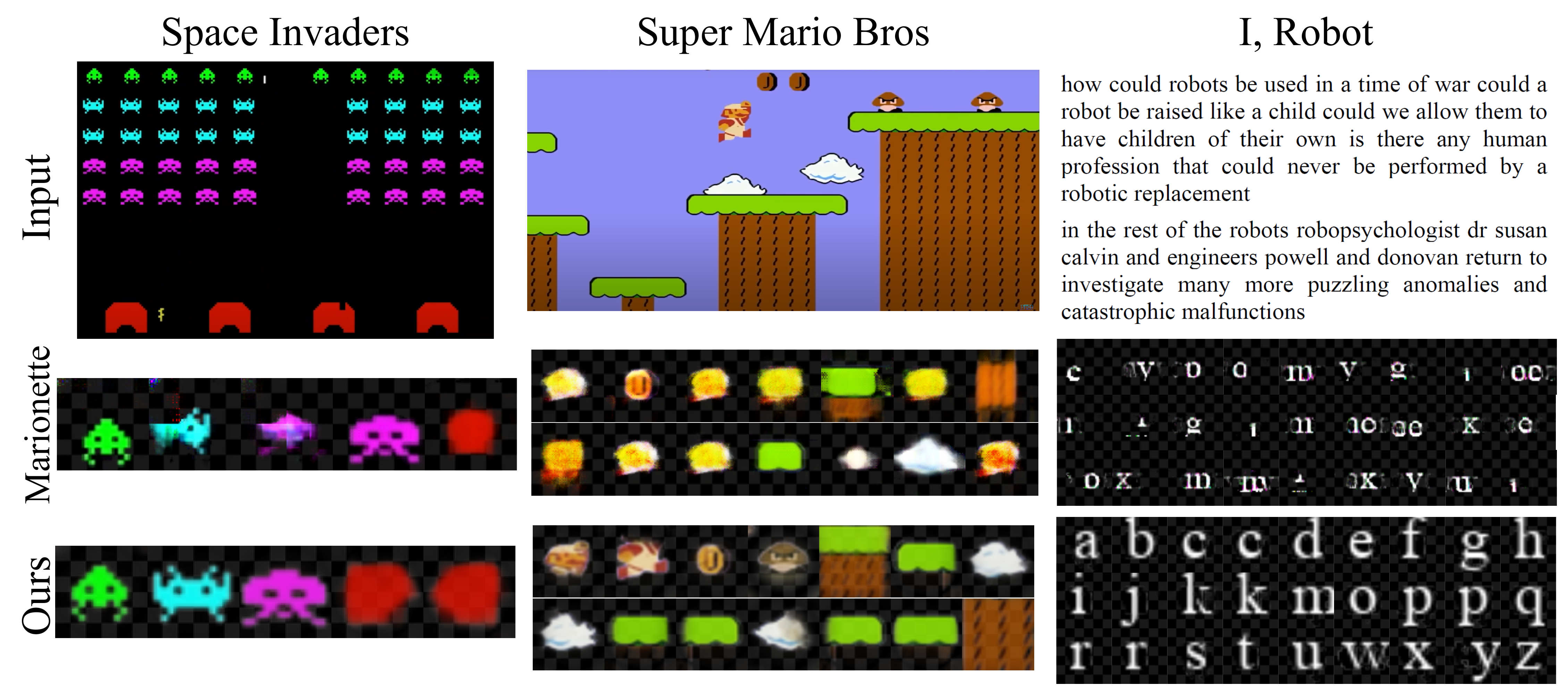}%
\vspace{5pt}
  \caption{\label{fig:mario} 
  Ours on MarioNette  and recovered concepts.
  %These results can be achieved using the code available in the supplementary.
  }
\vspace{-20pt}
\end{wrapfigure}
%\end{figure}
\begin{table}[t!]
% \caption{Segmentation and Cross dataset reconstruction results.}
\small
\centering
\caption{\label{tbl:ari} {\bf Segmentation performance.} Decomposition accuracy measured by the ARI metric.
Second best values are underlined, while 
%foreground metrics calculated between decomposition outputs of various methods and the ground truth segmentation.
($^*$) uses the same element transformations and number of concepts as our method.
%number of visual concepts and type of their transformations are homogenised to generate the results.
%($^{**}$) uses the discretized version of the dSprites Color dataset.
See Section \ref{datasets} for details.
%, second best are bold.
%that we report the ARI number on a discretized version dSprites Color dataset we explain the reason in detail in section \ref{datasets}. 
%M-dSprites refers to multi object dSprites dataset.
}
  \vspace{5pt}
  \begin{tabular}{r c c c c c}
    \toprule
  & Tetrominoes & dSprites Color. & dSprites Bin. & dSprites Adv. & CLEVR6 \\ 
  \midrule
    IODINE               & 99.2 $\pm$ 0.4 & 76.7 $\pm$ 5.6 & 64.8 $\pm$ 17.2 & - & \underline{98.8 $\pm$ 0.0}\\
    Slot Attention       & \underline{99.5 $\pm$ 0.2} & \underline{91.3 $\pm$ 0.3} & 69.4 $\pm$ 0.9 & 12.7 $\pm$ 1.1 & \textbf{98.8 $\pm$ 0.3}\\
    DTI-Sprites          & \textbf{99.6 $\pm$ 0.2} & \textbf{92.5 $\pm$ 0.3} & \underline{$75.5^* \pm$ 0.4} & \underline{$75.3^* \pm$ 0.4} & 97.2 $\pm$ 0.2\\
    Ours                 & \underline{99.5 $\pm$ 0.1} & $90.6 \pm$ 0.8 & \textbf{85.1 $\pm$ 0.7}  & \textbf{76.4 $\pm$ 2.4} & 64.6 $\pm$ 0.8 \\
    \bottomrule
  \end{tabular}
  \vspace{-12pt}
\end{table}

Figures~\ref{fig:tetris_and_clevr_results} and~\ref{fig:AB} show example decompositions on each dataset.
We provide the full dictionaries of visual concepts extracted from each dataset in the supplemental.
DTI-Sprites is most related to our method. Table~\ref{tbl:ari} shows our competitive segmentation performance, but Table~\ref{tbl:comparison} and Figure~\ref{fig:dti_vs_ours} reveal that the concepts it learns each may entangle multiple ground truth visual concepts, especially when using lower concept numbers. When reconstructing a scene, their image formation model needs to disentangle these concepts. Thus, it does often not correctly identify the concepts in a dataset.

In Fig~\ref{fig:mario} we should comparision between our method and Marionette, our method requires less data to extract concepts, only 4 and 6 frames for Space Invaders and Super Mario Bros, respectively, compared to several thousands of frames required by MarioNette.

\begin{table}[t]
% \subfloat[{\bf Cross dataset reconstruction.} Comparison of the MSE reconstruction loss on EMNIST letters for methods trained on MNIST digits. ]{
  \small
  \begin{minipage}[t]{0.45\textwidth}
  \centering

  \caption{{\bf Visual concept error.} Average $L_1$ distances between each library concept and the nearest learned concept.}
      \vspace{5pt}
  %decompositions and the nearest neighbour concept in dSprites dataset. We consider randomly sampled 1000 layers estimated by Slot Attention, learned DTI-Sprites sprites and visual concepts learned by our method as the decompositions.}
  \label{tbl:comparison}
  \setlength{\tabcolsep}{2pt} % General space between cols (6pt standard)
  \renewcommand{\arraystretch}{1} % General space between rows (1 standard)  
  \begin{tabular}{r c c}

    \toprule
  & dSprites Bin & dSprites Adv  \\ 
  \midrule
    Slot Att.                & 0.0312 & 0.0497   \\  
    DTI-Sprites                   & 0.0133 & 0.0219  \\ 
    Ours                          & \textbf{0.0033} & \textbf{0.0051} \\
    \bottomrule
  \end{tabular}
  \end{minipage}
%  \hfillx
\hspace{20pt}
  \begin{minipage}[t]{0.45\textwidth}
  \centering
  \caption{{\bf Cross-dataset reconstruction.} MSE reconstruction loss on EMNIST letters for methods trained on MNIST digits.}
  \vspace{3pt}
  \label{tbl:cross}
  \setlength{\tabcolsep}{2pt} % General space between cols (6pt standard)
  \renewcommand{\arraystretch}{1} % General space between rows (1 standard)
  \begin{tabular}{r c c }
    \toprule
  & MNIST(Train) &  EMNIST(Test)    \\ 
    % &      \\ 
  \midrule
    Slot Att.               & 0.0048   & 0.0560 \\  
    DTI-Sprites                & 0.0065   & 0.0202 \\
    Ours(128)                   & 0.0114  & 0.0169 \\
    Ours(512)                    & 0.0090   & 0.0140 \\
    \bottomrule
  \end{tabular}
  \end{minipage}
  \vspace{-10pt}
\end{table}

%\vspace{-10pt}
\subsection{Cross-dataset Reconstruction}
%\vspace{-5pt}
\begin{figure}[t]
% \begin{wrapfigure}[24]{rh}{0.6\textwidth}
  \centering
    % \vspace*{-.4in}
  {\includegraphics[width=\linewidth]{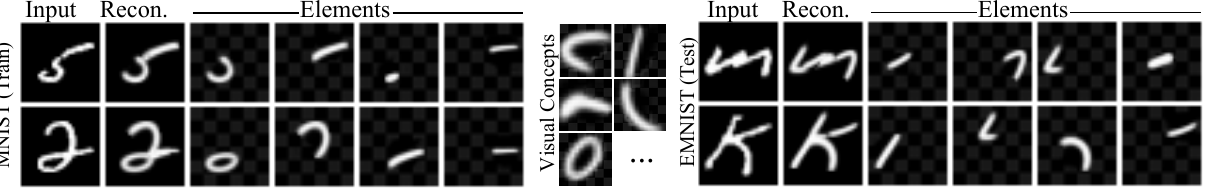}}%
\vspace{4pt}
  \caption{\label{fig:cross} The visual concepts learned on one dataset (MNIST, left), can generalize to a second dataset (EMNIST, right), without further training. 
%   Generalization. Qualitative results showing discovered decomposition on the  MNIST (training) dataset, resulting in a dictionary of strokes~(bottom). 
%   Keeping the dictionary fixed, ours can generalizes to explain a  different test (EMNIST) dataset. Notice the shared concept between letter m and letter k reconstruction.
  }

% \end{wrapfigure}
\vspace{-10pt}
\end{figure}

% We quantify cross dataset reconstruction using Root Mean Squared Error(RMSE).
% We also present nearest neighbour reconstruction on the decomposition's to quantify the decomposition performance.  

To quantify generalization performance, we train our algorithm and the baselines on the MNIST~\cite{lecun1998mnist} dataset, which contains hand-written digits, and test their reconstruction performance on EMNIST~\cite{cohen2017emnist} dataset, which also contains hand-written letters. Table~\ref{tbl:cross} shows a quantitative comparison between Slot Attention, DTI-Sprites and two versions of our method using the MSE reconstruction loss, with the visual dictionary size capped to $128$ or $512$. Figure~\ref{fig:cross} shows an example decomposition. Since we do not rely on learned priors in addition to our dictionary, our inference pipeline shows significantly better generalization performance than the baselines. Note that in this experiment, DTI-Sprites is allowed translation, rotation and scaling of elements, while our method only uses translations.

% \begin{table}[t!]
%   \centering
%   \caption{\label{tbl:cross}
%     {\bf New domain reconstruction.} ....
% }
%   \begin{tabular}{r c c c c}
%     \toprule
%   & Slot Attention & DTI-Sprites & Ours(128)  & Ours(512)  \\ 
%   \midrule
%     MNIST Train               & 0.0048  & 0 & 0.0114 & 0.0090\\  
%     EMNIST Test               & 0.0560  & 0 & 0.0175 & 0.0142 \\
%     \bottomrule
%   \end{tabular}
% \end{table}
\begin{figure}
\centering
\vspace{-5pt}
  \includegraphics[width=\linewidth]{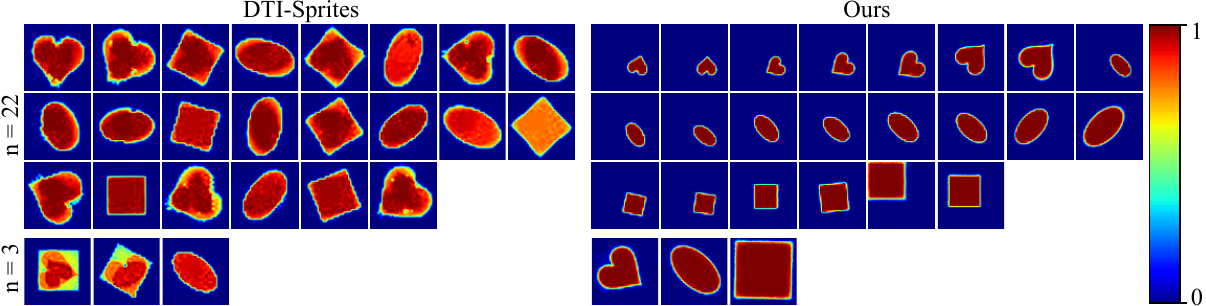}%
  \vspace{4pt}
  \caption{\label{fig:dti_vs_ours} Comparison between visual concepts learned by DTI-Sprites and our method on the M-dSprites Bin. dataset with different dictionary sizes $n$. The heatmap shows pixel intensities. DTI-sprites learns entangled concepts that correspond to multiple ground truths.% concepts, especially for small $n$.
  %the sprites learned using DTI-Sprites do not semantically correspond to the elements in dSprites dataset. Also notice that Our method learns elements at different scales as is expected since dSprites contains same object at multiple scales.
  }
  \vspace{-10pt}
\end{figure}

% \subsection{Decompostion reconstruction}
\iffalse
\subsection{3D Scene Reconstruction}

We create a synthetic dataset of $16$ 3D scenes by randomly selecting $3$ shapes from a dictionary of $4$ ground truth shapes and placing them at random positions on a ground plane. We render 20 different views of each scene to form the input dataset, and use the image formation described in Section~\ref{sec:3d_decomposition} to learn 3D concepts. Reconstruction results and a comparison of the learned concepts to the ground truth is provided in Figure~\ref{fig:3dsprites}.
% To test ``learning 3D concepts" on a  
% synthetic dataset with 16 scenes. Each scene contains 3 shapes randomly sampled from a set of 4 grouthtruth shapes. Each shape is placed at a random x,y coordinate on the ground i.e z=0. Each scene is the rendered from 20 different views and these renders form the input to the pipeline. 

% We compare the learned set of 3D concepts to the groundtruth shapes similar to table \ref{tbl:comparison}. The $L_1$ distance between each the learned 3D concepts and the nearest groundtruth shapes is 0.0088. We show a qualitative comparision in Figure \ref{fig:3dsprites}.
% \preddy{add fig of ground truth shapes and learned concepts Figure \ref{fig:3dsprites}}
\fi

%\vspace{-10pt}
\paragraph{Limitations.}
%\vspace{-5pt}
%Our pipeline is yet to be tested on a large scale real world dataset like ImageNet. 
Our pipeline has two main limitations.
%The major limitation to test on datasets like Imagenet is the computational cost of the search operation.
First, the computational cost of the element parameter search.
%limits the number of parameters we can use.
We plan to optimize the search operation using techniques like a coarse-to-fine search in future work.
Second, we currently search for exact repetitions of objects to learn our concepts.
%rather than accounting for deformations/variations. 
%This limits our ability to handle noise in the objects, for example due to low-resolution rasterization.
Accounting for deformations/variations by incorporating a more general parametric deformation model, for example by using a neural network with the help of differentiable rasterizers, will be a valuable next step towards a more general model. 
%rasterization noise, as shown on multi-dsprites, explicitly adding invariance to small variations, under a learnable deformation model. 

%\vspace{-10pt}
\section{Conclusion and Future Work}
%\vspace{-5pt}
We presented a general method to learn visual concepts from data, both for images and shapes, without explicit supervision or learned priors. Our main idea is posing the search for visual concepts as a direct optimization, which can be solved efficiently when splitting the task into alternating dictionary finding and parameter optimization steps. Using direct optimization,  instead of a network-based approach,  improves the quality of the resulting visual concepts and additionally reveals parameters such as hue, position, and scale that are not available to most network-based approaches.
%results in a search-and-optimize paradigm that discovers interpretable elements, that also helps to generalize to new (test) data. 
In the future, we would like to extend our approach to a fully generative model.
%additionally learn a generative model directly on the input data.
One approach would be to
%use the extracted decompositions and
learn a distribution over the element parameters. When combined with the learned concepts, we could sample the element parameter distributions to produce new images with the given image formation function.
%and then combine that with the learned elements to directly produce images, via the given compositing functions.
%Interestingly, such an approach can be optimized end-to-end via a differentiable rasterization module~\cite{reddy2020discovering}. 
This opens up new avenues for parametric generative models, blurring the line between neuro-symbolic and image-based generative models. 
We believe that ultimately the right direction for a decomposition is a hybrid between network-based and search-based methods. %In this work, we propose an antithesis to the current purely network-based approaches and hope to inspire a combination of both in the future. 
%we aim to add priors to the search operation for computational efficiency and invariance to certain deformations. We also aim to extend our method to more modalities. One interesting future direction would be to couple our decomposition pipeline with neural network methods. 

% \pagebreak

%\paragraph{Potential negative societal impact} at the current stage of the algorithm we do not see any potential negative societal impacts. 

\if0
\section*{Acknowledgments}

We would like to thank the anonymous reviewers for their helpful suggestions and Maks Ovsjanikov for discussions in an early phase of this project. Rabin Ezra Scholarship.
This research was supported by an ERC Grant (SmartGeometry 335373), Google Faculty Award, and gifts from Adobe.
\fi

\bibliography{macros,main}

\clearpage

\newcommand{\beginsupplement}{%
        \setcounter{section}{0}
        \renewcommand{\thesection}{S.\arabic{section}}%
        \setcounter{table}{0}
        \renewcommand{\thetable}{S\arabic{table}}%
        \setcounter{figure}{0}
        \renewcommand{\thefigure}{S\arabic{figure}}%
        \setcounter{equation}{0}
        \renewcommand{\theequation}{S\arabic{equation}}
     }

\beginsupplement
% Still needed:
% - More 3D results  ---Todo
% - Street signs dataset -- Partial (should we also show decompositions?)
% - More examples on patterns (do we have other large datasets of patterns? Can we overfit on a single pattern image?)

% Done:
% - ARI calculation
% - More 2D results
% - full visual concepts for each dataset   --- Done
% - background handling?                    --- Done
% - Detailed derivation of the multi-layer convolutions (Eq. 7)  -- Done
% - More dataset details (especially dSprites adversarial) --- Done
% - Grid search resolution for each layer parameter  ---Done
% - Optimization video - Done
% - Different composition functions ---Done
% - Qualitative Ablation: With vs. without visual concept cloning figure - Done
% - Ablation: comparison of optimizers - Done
% - Code (can also be later) - Done

\section{Convolution-based Grid Search}
When optimizing the element parameters in a given image $\tilde{I}$, we maximize the normalized cross-correlation~\cite{10.1371/journal.pone.0203434} instead of minimizing the $L_2$ distance:
% \vspace{-5pt}
\begin{equation}
    \argmax_{\theta_1, \dots, \theta_n} \frac{\sum_{p} (I \tilde{I})_p}{\sqrt{\sum_{p}I^2_p} \hspace{5pt} \sqrt{\sum_{p} \tilde{I}^2_p}} \label{eq:correlation} %\\ \nonumber
  \quad   \text{with } \tilde{I} = h(e_\mathcal{V}(\theta_1), \dots, e_\mathcal{V}(\theta_n)),
\end{equation}
where the products are element-wise and p is a two-dimensional pixel index.
We can then formulate a grid search over translation parameters as a convolution. In the single-element case, $\tilde{I} = E_1$, and we can rewrite Eq.~\ref{eq:correlation} as:
% l_\mathcal{V}(\theta_1^{t=0}
% \vspace{-12pt}
\begin{equation}
    \argmax_{\theta_1} \frac{(I \circledast \hat{E}_1)_{t_1}}{\sqrt{\sum_{p}I^2_p} \hspace{5pt} 
    \sqrt{(\mathbf{1} \circledast \hat{E}_1^2)_{t_1}}},
    \label{eq:convolution}
\end{equation}
where $\hat{E}$ is the element $E$ with translation parameters $t \in \theta$ zeroed out, so the transformed visual concept is at the origin and acts as a convolution kernel, and $\mathbf{1}$ is an image of all-ones. The result of the convolution is an image where each pixel corresponds to the correlation at one 2D translation $t$.

In the general case with multiple elements, we optimize the parameters of one element at a time. At each element, we need to account for occlusions from previous elements when performing the convolution. Re-writing the alpha-composite defined in Eq. 1 to isolate the contribution of a single element $E_i$ we get:
% \vspace{-12pt}
\begin{gather}
    \tilde{I} = \tilde{I}_1^{i-1} + E_i O_1^{i-1} + (1-E_i^{\text{A}}) O_1^{i-1} \tilde{I}_{i+1}^{n}\\ 
    \text{with   }    \tilde{I}_a^b \coloneqq \sum_{i=a}^b E_i O_a^{i-1} \; 
    \text{and } O_a^j \coloneqq \prod_{i=a}^j (1 - E_i^{\text{A}}).\nonumber
\end{gather}
Intuitively, $\tilde{I}_a^b$ is the partial composite of the elements $a$ through $b$, without taking into account other elements, and $O_a^j$ is the occlusion effect of elements $a$ through $j$ on the following elements. Only the second and third terms depend on $L_i$, where the second term is the occluded contribution of $L_i$ to the image and the third term describes the occlusion caused by $L_i$ on the following elements. Substituting into Eq.~\ref{eq:correlation}, and using convolutions for searching over translations we arrive at:
\begin{gather}
\argmax_{\theta_i} \frac{\sum_p (IC_1)_p + (IC_2 \circledast \hat{E}_i + IC_3 \circledast (1-\hat{E}_i^A))_{t_i}}{\sqrt{\sum_p I^2_p} \hspace{5pt}  \sqrt{\sum_p (C_1^2)_p + (\mathcal{N}_i)_{t_i}}}\\ \nonumber
\begin{aligned}
\text{with } \mathcal{N}_i =\ &C_2^2 \circledast \hat{E}_i^2 + C_3^2 \circledast (1-\hat{E}_i^A)^2 \\
 &+ 2 C_1 C_2 \circledast \hat{E}_i + 2 C_1 C_3 \circledast (1-\hat{E}_i^A)\\
 &+ 2 C_2 C_3 \circledast \hat{E}_i (1-\hat{E}_i^A)
\nonumber
\end{aligned}\\
\text{and } C_1 = \tilde{I}_1^{i-1} \hspace{10pt} C_2 = O_1^{i-1} \hspace{10pt} C_3 =  O_1^{i-1} \tilde{I}_{i+1}^{n}, \nonumber
\end{gather}
which we solve as efficient update step for element parameters. Section~\ref{sec:layer_param} provides the derivation.

\section{Layer Parameter Objective}
\label{sec:layer_param}
We derive the objective for the layer parameter optimization with alpha compositing and convolutions defined in Eq. 7 of the main paper. The objective is obtained in two steps: (i) we substitute the alpha composite defined in Eq. 6 into layer parameter objective in Eq. 4, and (ii) we search over position parameters using convolutions, analogous to Eq. 5. For clarity, we first define:
\begin{equation}
    C_1 = \tilde{I}_1^{i-1} \hspace{10pt} C_2 = O_1^{i-1} \hspace{10pt} C_3 =  O_1^{i-1} \tilde{I}_{i+1}^{n}, \nonumber
\end{equation}
so that Eq. 6 becomes:
\begin{equation}
    \tilde{I} = C_1 + E_i C_2 + (1-E_i^{\text{A}}) C_3.
    \label{eq:alpha_comp}
\end{equation}
Substituting Eq.~\ref{eq:alpha_comp} into Eq. 4 and restricting the optimization over the parameters $\theta_i$ of layer $E_i$:
\begin{gather}
\argmax_{\theta_i} \frac{\sum_p (IC_1)_p + \sum_p(IC_2 E_i)_p + \sum_p(IC_3 (1-E_i^A))_p}{\sqrt{\sum_p I^2_p} \hspace{5pt}  \sqrt{\sum_p (C_1^2)_p + \mathcal{N}'_i}}\\ \nonumber
\begin{aligned}
\text{with } \mathcal{N}'_i =\ & \textstyle\sum_p(C_2^2 E_i^2)_p\\
& + \textstyle\sum_p(C_3^2 (1-E_i^A)^2)_p\\ 
& + 2 \textstyle\sum_p(C_1 C_2 E_i)_p\\
& + 2 \textstyle\sum_p(C_1 C_3 (1-E_i^A))_p\\
& + 2 \textstyle\sum_p(C_2 C_3 E_i (1-E_i^A))_p.
\nonumber
\end{aligned}
\end{gather}
Finally, exchanging any term of the form $\sum_p(C E_i)_p$ with a convolution $(C \circledast \hat{E}_i)_{t_i}$ using as kernel the layer with with zeroed translation parameters $\hat{E}_i$, we arrive at Eq. 7 of the paper:
\begin{gather}
\argmax_{\theta_i} \frac{\sum_p (IC_1)_p + (IC_2 \circledast \hat{E}_i + IC_3 \circledast (1-\hat{E}_i^A))_{t_i}}{\sqrt{\sum_p I^2_p} \hspace{5pt}  \sqrt{\sum_p (C_1^2)_p + (\mathcal{N}_i)_{t_i}}}\nonumber\\ \nonumber
\begin{aligned}
\text{with } \mathcal{N}_i =\ &C_2^2 \circledast \hat{E}_i^2\\
& + \sum C_3^2 - 2C_3^2 \circledast  \hat{E}_i^A  + C_3^2 \circledast (\hat{E}_i^{A})^2\\ 
& + 2 C_1 C_2 \circledast \hat{E}_i\\
& + 2 \sum C_1 C_3  - C_1 C_3 \circledast \hat{E}_i^A\\
& + 2 C_2 C_3 \circledast \hat{E}_i - 2 C_2 C_3 \circledast \hat{E}_i \hat{E}_i^A.
\nonumber
\end{aligned}
\end{gather}

\section{Datasets and Hyperparameters details}
\label{Sec:Hyperparameters}
The Tetrominoes dataset contains $60$k images with 3 randomly rotated and positioned Tetris blocks. 
The Multi-dSprites dataset consists of $60$k images, each showing 2-3 shapes picked from a dictionary and placed at random locations, possibly with occlusions. We use a discretized variant of the color version and the binarized version of this dataset. We discretize the color version to eight colors, so that the colors can be discovered as composite concepts. In addition, we create a variant of this dataset that we name \emph{Multi-dSprites adversarial}, where shapes can only appear in three discrete locations on the canvas. This intuitively simple dataset highlights shortcomings in existing methods.
% We show that this dataset is difficult to handle for some existing methods.
The Clevr6 dataset consists of $35$k rendered 3D scenes with each scene is an arrangement of up to six geometric primitives with various colors and materials. Also since MarioNette data is not available, we obtain comparable data through screen captures of \href{https://www.youtube.com/watch?v=aj43i9Az4PY}{Space Invaders}, \href{https://www.youtube.com/watch?v=rLl9XBg7wSs}{Super Mario Bros}, and \href{https://archive.org/details/irobotnovel/page/n7/mode/2up}{I, Robot} (in the latter we use only lower-case letters). 

\paragraph{Optimization details.}
%\vspace{-10pt}
%In all our experiments
We learn visual concepts using AdaDelta~\cite{zeiler2012adadelta} with a learning-rate of $1.0$ and a batch size of $8$ on a single GPU. 
We have tested different optimizers but found that our pipeline was robust to the choice of optimizer. A more detailed comparison of optimizers is  in the supplemental.
%a graph of mse and the choice of optimizers in the supplementary.
We do not use any learning rate schedulers or warm-up techniques.

In Table~\ref{tbl:hyperparameters} we list the hyperparameters we used for each experiment. %of the experiments we reported numbers of in the main paper and the supplementary.
\begin{table*}
  \small
  \centering
  \caption{\label{tbl:hyperparameters}
    {\bf Hyperparameters.} In this table present the hyperparameters used of different experiments that are used to generate the results presented in the paper. Translation values are greater than the input image size if we allow partial placement of a concept. The first and second columns show the initial number of concepts and the max number of concepts . The rotation column indicates the number of values uniformly sampled in $[0, 2\pi]$.   
}
\vspace{3pt}
s
  \begin{tabular}{r c c c c c c c c}
    \toprule
  & Init  &  Max  & Translation  & Rotation & nColors & nlayers & Concept Size    \\ 
  \midrule
    Tetris                    & 1  & 10 &  35x35 & 4 & 6 & 3 & 19\\
    Multi dSprites bin        & 4   & 22 & 64x64 & 40 & 1 & 3 & 33\\
    Multi dSprites Adv        & 1   & 3 & 64x64 & 40 & 1  & 3 & 33\\
    Clevr6                    & 8   & 32  & 64x64 & 1 & 8 & 6 & 31\\
    MNIST(128)                & 8   & 128 & 28x28 & 1 & 1 & 4 & 13\\
    MNIST(512)                & 8   & 512 & 28x28 & 1 & 1 & 4 & 13\\
    MNIST Sum(128)            & 8   & 128 & 41x41 & 1 & 1 & 4 & 13\\
    MNIST Sum(512)            & 8   & 512 & 41x41 & 1 & 1 & 4 & 13\\
    GTSRB            & 6   & 64 & 28x28 & 1 & 1 & 6 & 31\\
    Xmas Pattern            & 2   & 2 & 85x85 & 1 & 1 & 45 & 21\\
    \bottomrule
  \end{tabular}
\end{table*}

\section{Thresholds for Splitting and Removing Concepts}
The threshold for removing a concept $V_j$ is $N_j < 0.25 \frac{n|\mathcal{I}|}{|\mathcal{V}|}$, where $N_j$ is the number of instances of concept $v_j$, and $n$ is the number of elements per image. The threshold for splitting a concept is $N_j > 0.25 \frac{n|\mathcal{I}|}{|\mathcal{V}|}$, in addition to a threshold on the reconstruction error $\mathcal{E}_{j} < 0.95$. The reconstruction error $\mathcal{E}_{j}$ of a single visual concept $V_j$ over the whole dataset is isolated as:
\vspace{-5pt}
\begin{equation}
    \mathcal{E}_{j}(\mathcal{V}, \Theta) \coloneqq \sum_{\{i, k | \tau^k_i = j\}} \frac{\sum_p (M^k_i I \tilde{I} M^k_i )_p}{\sqrt{\sum_p (M^k_i I)_p^2} \sqrt{\sum_p (M^k_i \tilde{I})_p^2}} %\\
    \quad \text{with } M^k_i = (E^A_i)^k (O_1^{i-1})^k. % \nonumber.
\end{equation}
The sum is over all elements $E^k_i$ in all images $I^k$ that use the visual concept $V_j$ and $M^k_i$ is a mask that zeros out parts of the image that do not have contributions from element $E^k_i$.

\section{Concept Evolution Graph}
\begin{figure*}[t]
\centering
  \includegraphics[width=\textwidth]{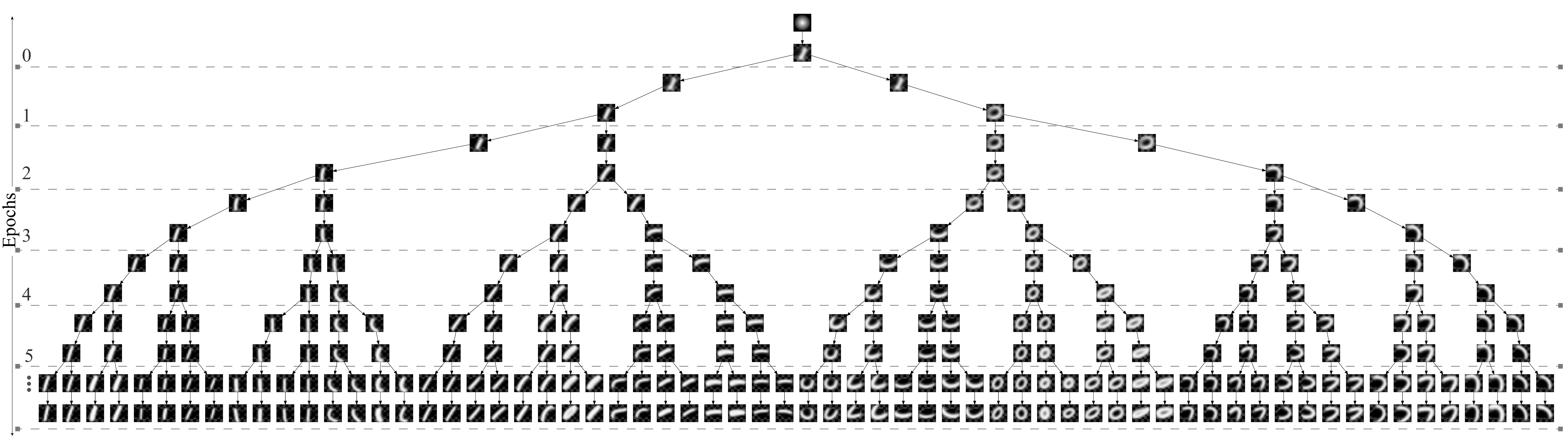}%
  \vspace{5pt}
  \caption{\label{fig:cloneGraph} 
  Visual concept evolution for the MNIST dataset. Note how the concepts evolve over optimization epochs by specializing to subtle stroke variations. 
  }
  \vspace{-15pt}
\end{figure*}

As described in the main paper, the number of concepts is determined by a concept evolution approach where concepts can be split or removed.
This results in a tree of visual concepts that is grown during optimization. An illustration of such a tree is shown in Figure~\ref{fig:cloneGraph}.

\section{Learning Visual Concepts in 3D Scenes}
\label{sec:3d_decomposition}
We use our framework to learn 3D visual concepts $\mathcal{V}$ from multi-view renders of 3D scenes. In this setting, our visual concepts are 3D voxel patches instead of 2D image patches, where each voxel describes a density value.
The transformation function $T$ translates these patches and samples them at the global voxel grid of the 3D scene to obtain 3D elements $E_i$:
%\begin{equation}
    $E_i = T(V_{\tau_i}, t_i),$
%\end{equation}
where $t_i$ is a 3D translation and $V_{\tau_i}$, a visual concept, is a 3D voxel patch containing density values.
For the image formation function $h$, we use the orthographic projection to accumulate the voxel densities along the viewing direction $d$, giving us the image $\tilde{I}$:
\begin{equation}
\tilde{I} = h(E_1, \dots, E_n | d) \coloneqq \sum_l \big(S^2_{jkl} \prod_{m=1}^{l-1} (1-S_{jkm})\big), \text{ with } S = \min(1,R_d\big(\sum_i{E_i}\big)),
\end{equation}
where $S$ is the scene voxel grid, computed as a sum over all element voxel grids, rotated and re-sampled by $R_d$ to align the last axis (indexed by $l$) with the viewing direction $d$. We clamp densities in $S$ to have a maximum of value $1$. The orthographic projection accumulates voxel densities along the viewing direction and the product attenuates voxel contributions by the occlusion effect of voxels closer to the viewer, indicated by smaller index $l$.
We optimize element parameters $t_i$ and visual concepts $\mathcal{V}$ to maximize the normalized cross-correlation between all reconstructed renders $\tilde{I}$ and all ground truth renders $I$ of a 3D scene dataset, as described in the previous sections, but without using convolutions to speed up the search for element parameters. The element parameters $t_i$ of a scene are optimized using all views of that scene. In our experiments, we use $20$ views: back/front, left/right, top, and 3 additional rotations of these views by ${\pi}/{8}$ radians about the top/bottom axis.
\begin{figure*}[t]
  \centering
  \subfloat
  {\includegraphics[width=7.0cm]{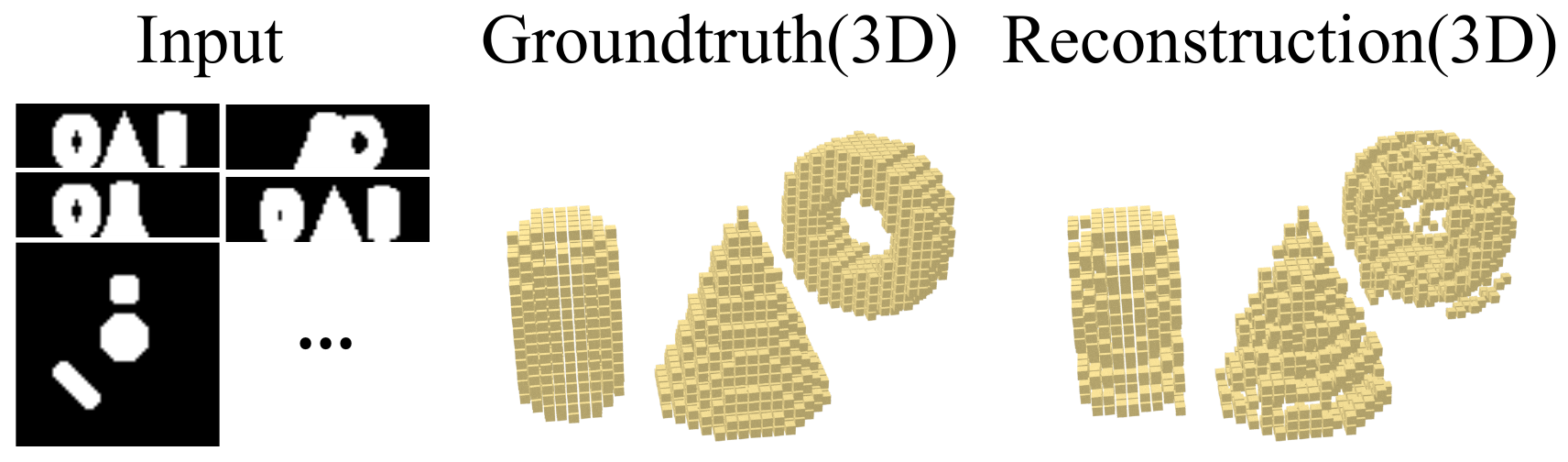} \label{fig:tetris_results}}
  \qquad
  \subfloat
  {\includegraphics[width=3.5cm]{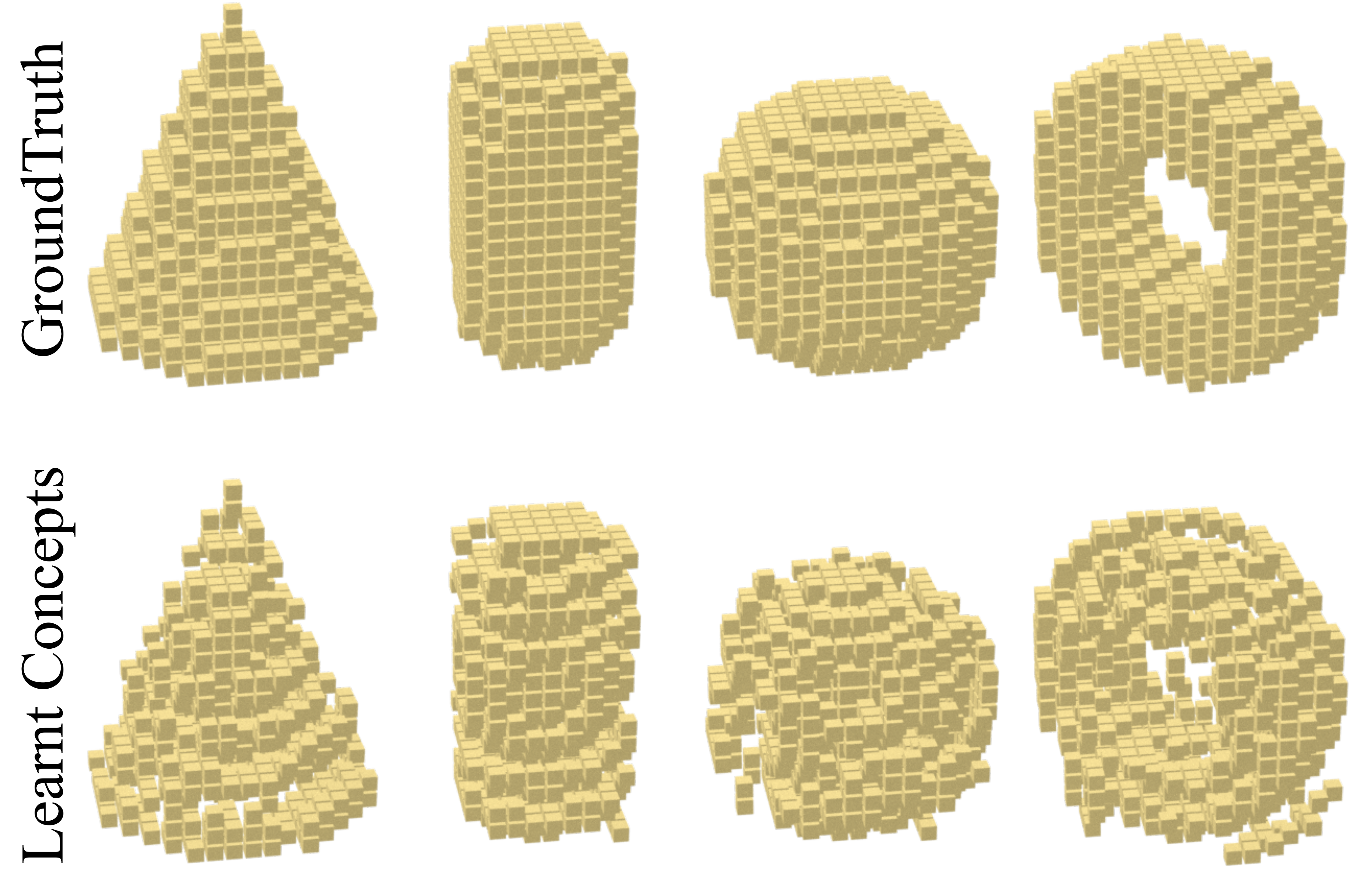} \label{fig:clevr_results}}
\vspace{3pt}
\caption{Given multi-view 2D renders of 3D scenes (left), our method can learn 3D visual concepts that can be used to reconstruct the 3D scenes.}
\label{fig:3dsprites}
\vspace{-15pt}
\end{figure*}

\paragraph{Evaluation:}We create a synthetic dataset of $16$ 3D scenes by randomly selecting $3$ shapes from a dictionary of $4$ ground truth shapes and placing them at random positions on a ground plane. We render 20 different views of each scene to form the input dataset, and use the image formation described in Section~\ref{sec:3d_decomposition} to learn 3D concepts. Reconstruction results and a comparison of the learned concepts to the ground truth is provided in Figure~\ref{fig:3dsprites}.

\section{Additive Compositing}
\begin{figure*}[t]
%\begin{wrapfigure}[16]{rh}{0.5\textwidth}
\centering
 %\vspace*{-.3in}
  \includegraphics[width=0.5\linewidth]{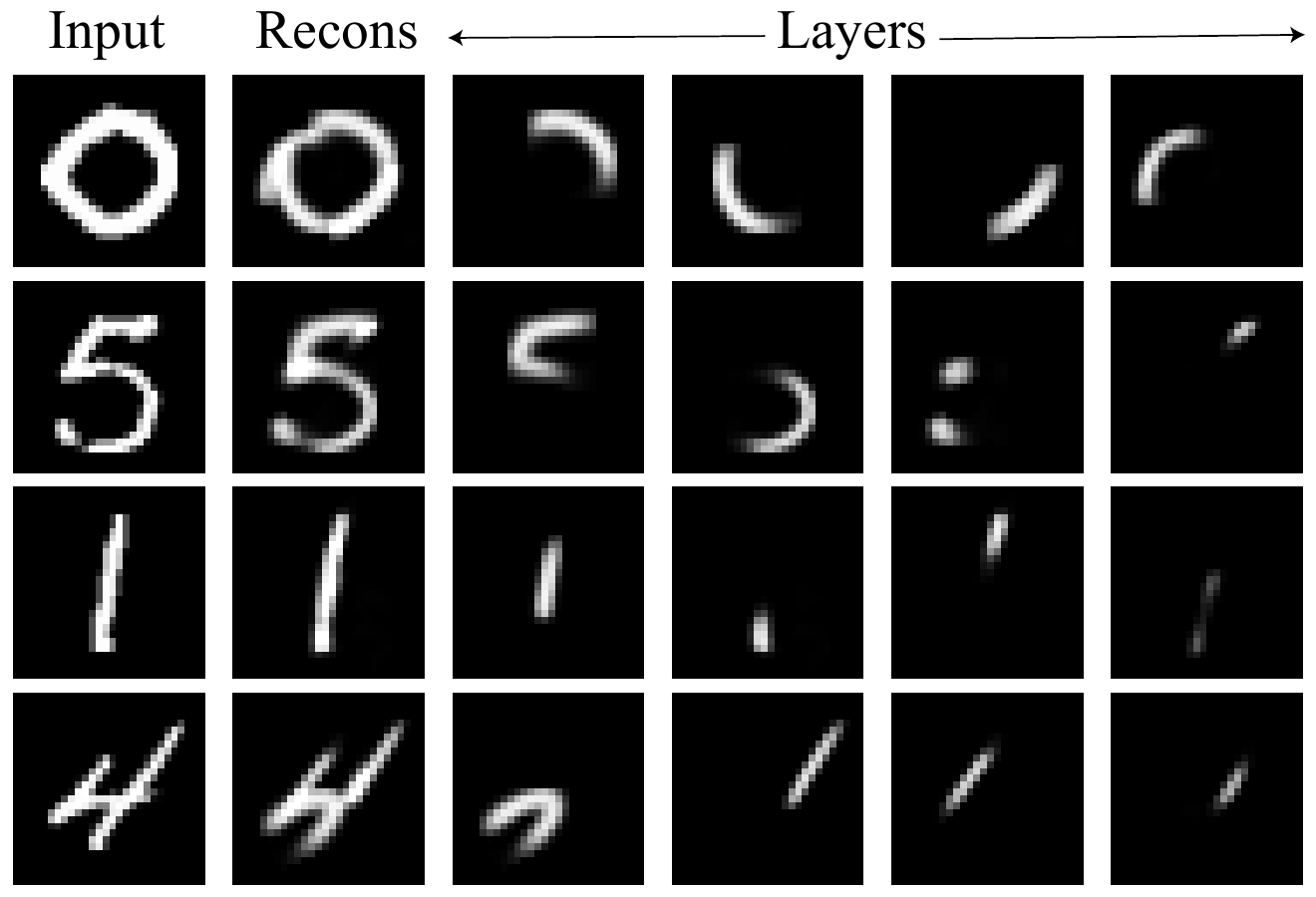}%
  \vspace{3pt}
\caption{\label{fig:mnist_sum_recons} 
  Example decomposition of an MNIST digit using additive compositing. 
  }
%\end{wrapfigure}
\end{figure*}

As mentioned in the main paper, we can use our approach with different compositing functions. Here, we present details and experiments with \emph{additive compositing} instead of the alpha-compositing used for the 2D results in the main paper and the other sections of the supplementary. For additive compositing, we replace Eq. 6 with a sum over layers:
% Here we present details of our pipeline with an additive compositing functions. We use additive compositing for 
% We using summation compositing function like eq.  \ref{eq:alpha_comp_summation} for this experiment. We present the cross dataset reconstruction results in table. \ref{tbl:comparison}. In Fig. \ref{fig:mnist_sum_graph} and \ref{fig:mnist_sum_recons} we present the clone graph and layer decomposition respectively. We also present the full set of concepts learned by training with summation compositing in Fig. \ref{fig:mnist_sum_concepts}.
\begin{gather}
    \tilde{I} = C + E_i \\
    \text{with } C = \sum_{j\neq i} E_j. \nonumber
\end{gather}
% \begin{equation}
%     \tilde{I} = C + L_i.
%     \label{eq:alpha_comp_summation}
% \end{equation}
The layer parameter optimization objective defined by Eq. 7 for alpha composting then becomes the following for additive compositing:
% Exchanging terms similar to eq. \ref{eq:alpha_comp} like shown in sec. \ref{sec:layer_param}, we arrive at:
\begin{gather}
\argmax_{\theta_i} \frac{\sum_p (IC)_p + (I \circledast \hat{E}_i)_{t_i}}{\sqrt{\sum_p I^2_p} \hspace{5pt}  \sqrt{\sum_p (C^2)_p + (\mathbf{1} \circledast \hat{E}^2_i)_{t_i} + (2 C \circledast \hat{E}_i)_{t_i}}}. %(\mathcal{N}_i)_{t_i}}}\nonumber\\ \nonumber
% \begin{aligned}
% \text{with } \mathcal{N}_i =\ &C_2^2 \circledast \hat{L}_i^2\\
% & + 2 C_1 C_2 \circledast \hat{L}_i\\
% \nonumber
% \end{aligned}
\end{gather}
\begin{figure*}
\centering
  \includegraphics[width=\textwidth]{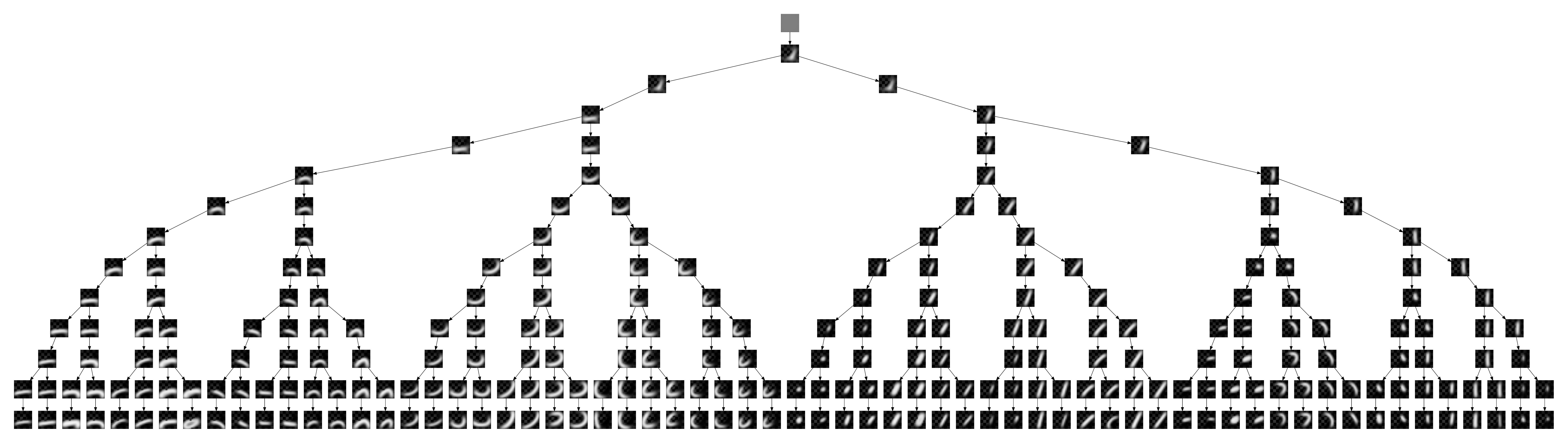}%
  \caption{\label{fig:mnist_sum_graph} 
\vspace{5pt}
  Graph illustrating visual concept evolution for the MNIST dataset trained using additive compositing. (Please use digital zoom for details.) Note how the method learns smaller concepts compared to alpha compositing, as shown in Figure~4 of the main paper. 
  }
\end{figure*}
Quantiative results measuring the MSE reconstruction loss for the cross-dataset generalization experiment are shown in Table~\ref{tbl:comparison} for two dictionary sizes. Note that reconstruction errors are slightly higher with additive compositing when compared to the corresponding results with alpha-compositing in Table~3 of the main paper. Figures~\ref{fig:mnist_sum_concepts}  and~\ref{fig:mnist_sum_recons} show the learned concepts and a decomposition example, respectively. Figure~\ref{fig:mnist_sum_graph} show the evolution graph of the visual concepts.
\begin{figure}
\centering
  \includegraphics[width=\linewidth]{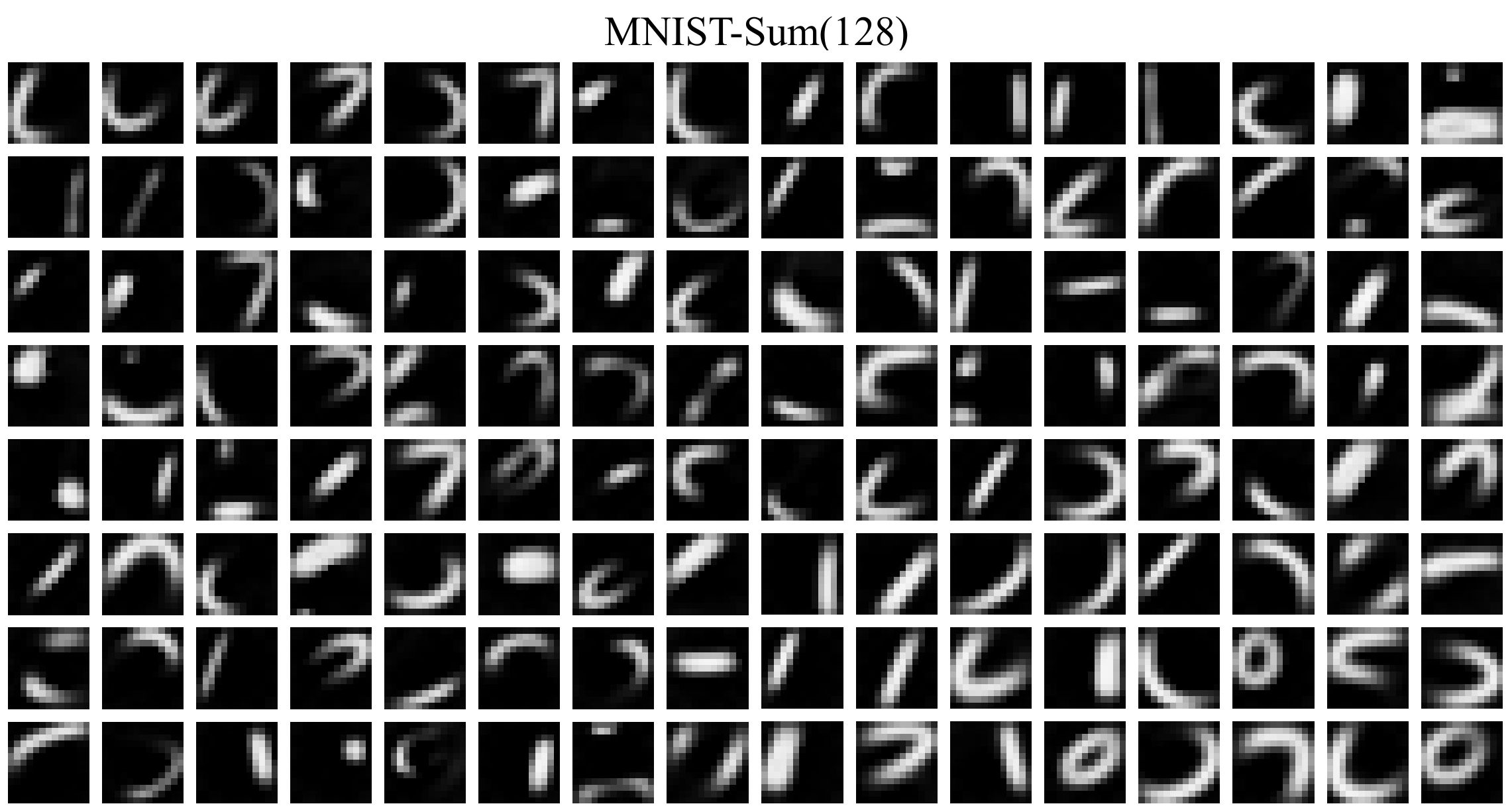}%
  \vspace{3pt}
\caption{\label{fig:mnist_sum_concepts} 
  Visual concepts learned from the MNIST dataset using additive compositing. 
  }
\end{figure}

\begin{table}[t!]
  \small
  \centering
  \caption{\label{tbl:comparison}
    {\bf Additive compositing results.} MSE reconstruction loss of EMNIST letters using additive compositing. The visual concepts are trained on the MNIST digits dataset with an additive compositing function. We show results for two dictionary sizes, $m=128$ and $m=512$.
}
  \begin{tabular}{r c c }
    \toprule
  & MNIST(Train) &  EMNIST(Test)    \\ 
  \midrule
    Ours Additive (128)                   & 0.0163  & 0.0215 \\
    Ours Additive (512)                    & 0.0137   & 0.0186 \\
    \bottomrule
  \end{tabular}
\end{table}

\section{Additional Results}
We also submit an additional set of uncurated results along with the supplementary (see the contents of the zip file). We included the first $b$ images of the respective datasets, with $b$ being the batch-size. For each image, we show the reconstruction and the decomposed layers. Note that these results are not post-processed, so the layer decomposition may also contain layers that are completely occluded by other layers.

\section{Comparison with Traditional methods}
\begin{figure}
%\begin{wrapfigure}[16]{1.0\textwidth}
\centering
    % \vspace*{-.35in}
  \includegraphics[width=\linewidth]{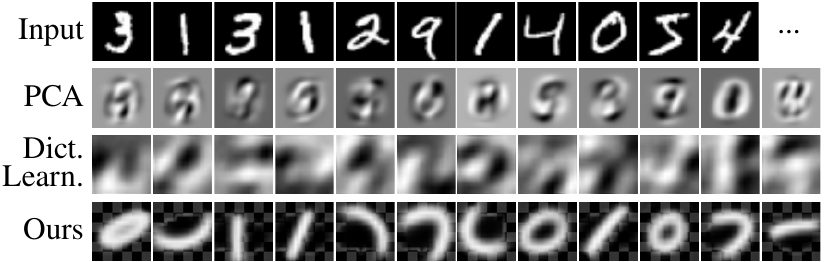}%
\vspace{3pt}
  \caption{\label{fig:pattern_oldschool_decomp} 
  We show the performance of classical methods like PCA or dictionary learning to obtain dominant modes of an image dataset. Note that extracted modes are not very interpretable in terms of compositionality. 
  In contrast, our method uses a search-and-learn strategy to extract a dictionary of interpretable \emph{visual concepts}, in this case the underlying strokes. Checkered patterns denote transparent pixels. }
%\end{wrapfigure}
\end{figure}

\begin{figure}[b!]
  \centering
  \vspace*{-25pt}
  \subfloat%[a][Decomposition results on the binarized M-dSprites dataset. Slot Attention is not regularized by a global dictionary, resulting in an incorrect decomposition.
  %Layer decompostion comparision between Our inference optimization and Slot Attention on M-dSprites bin dataset, since Slot is not constrained it estimates shapes that are not part of dSprites dataset.]
  {\includegraphics[width=5.5cm]{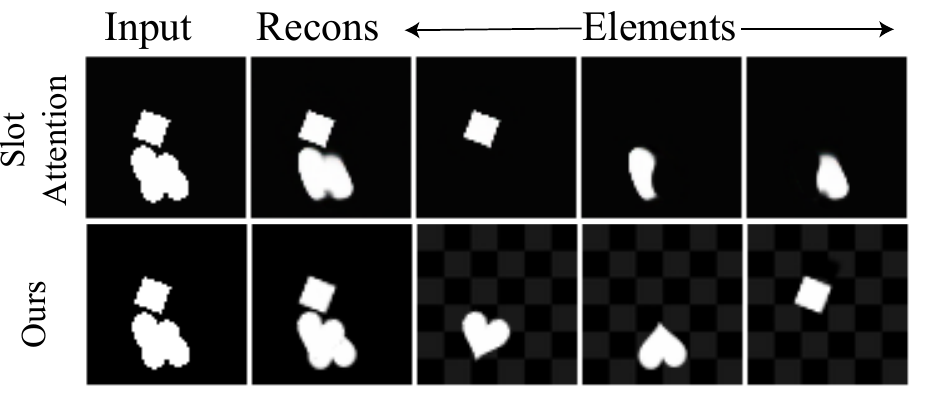} \label{fig:a}}
  \qquad
  \subfloat%[b][Decomposition results on the adversarial M-dSprites dataset. Slot attention fails on this intuitively simpler version of the dataset.]
  {\includegraphics[width=5.5cm]{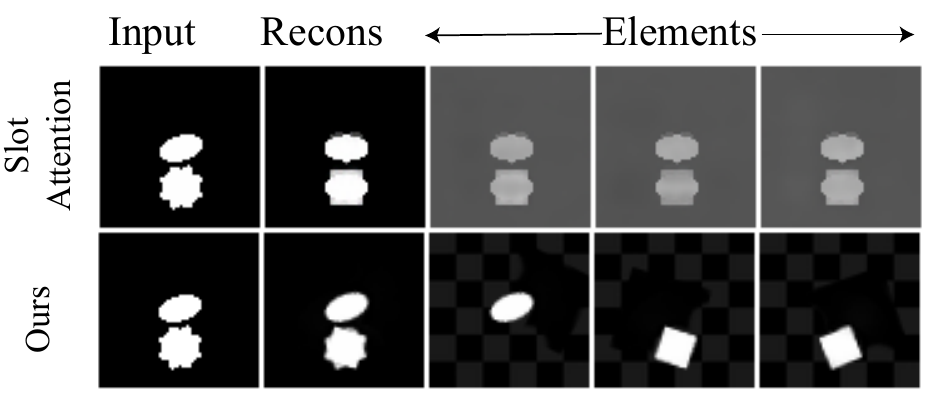} \label{fig:b}}
\vspace{3pt}
\caption{Comparison of decomposition result to Slot Attention on the M-dSprites Bin. and M-dSprites Adv. datasets. Slot Attention is not regularized by a global dictionary, resulting in incorrect decompositions. Ours are more interpretable. } \label{fig:AB}
\vspace{-15pt}
\end{figure}

We compare our approach to traditional unsupervised decomposition methods like PCA or dictionary learning in Figure~\ref{fig:pattern_oldschool_decomp}.

\section{Ablation of Optimizers}
\begin{figure}[h!]
\centering
    \vspace*{-.5in}
  \includegraphics[width=0.6\textwidth]{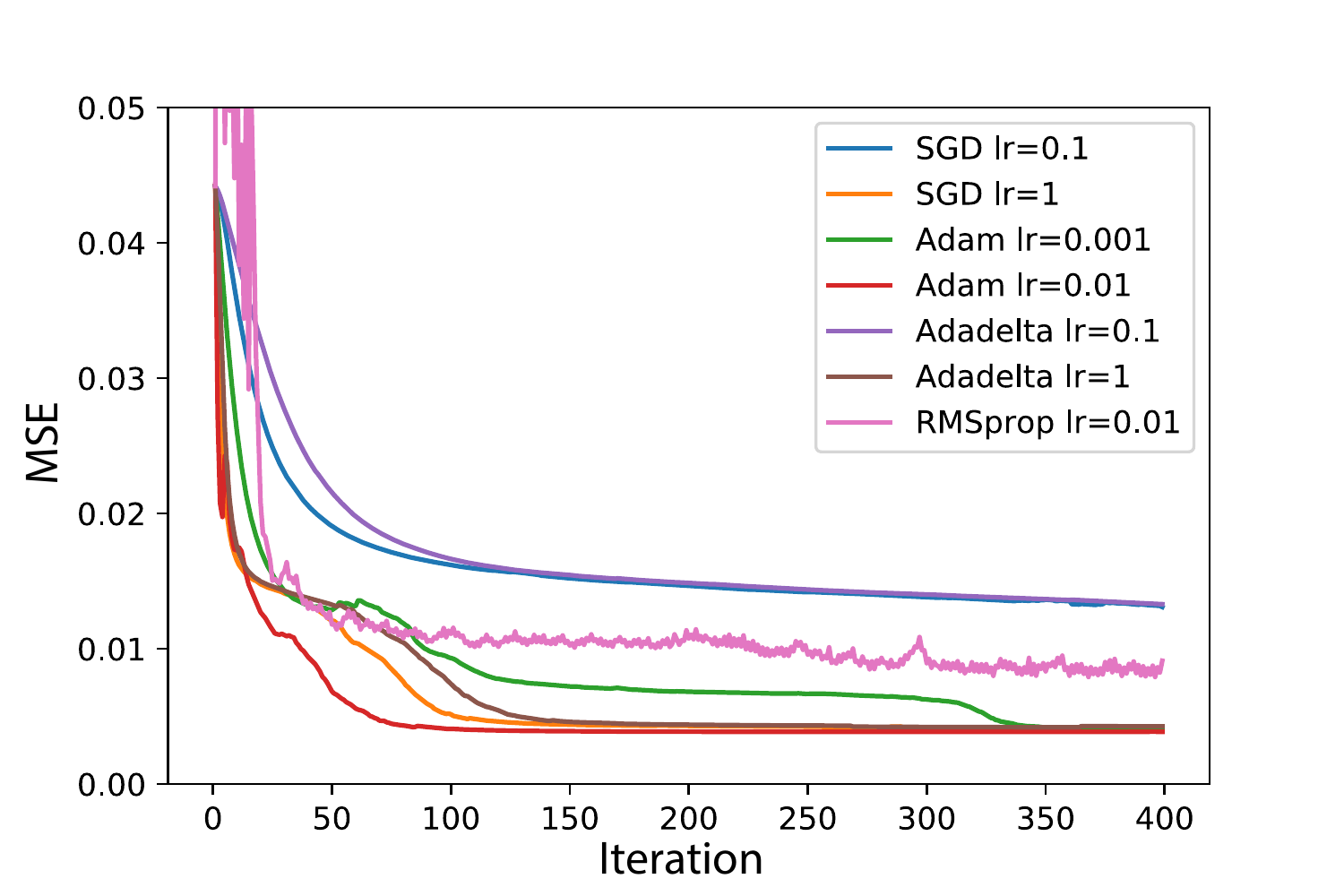}%
  \vspace{3pt}
\caption{\label{fig:plot} MSE training loss curves for various optimizers and learning rates. Note that all optimizers converge with the right learning rate.
  }
\end{figure}

% A sign of a well designed technique is robustness to choice of optimizers and choice of optimization parameters.
We demonstrate the robustness of our approach to the choice of optimizer and learning rates.
In Figure~\ref{fig:plot}, we show the MSE training loss curve for different optimizers and learning rates on the pattern images shared along with this supplementary material.
%vs Iteration curve for the pattern images shared along with the supplementary.
We show SGD, Adam \cite{kingma2014adam}, Adadelta \cite{zeiler2012adadelta}, and RMSprop optimizers with learning rates in $[0.001, 1]$. Our pipeline converges for all the optimizers with appropriate learning rate but lower learning rates take longer to converge.
%. In instances with lower learning rate the pipeline takes longer to converge which is expected.  

\section{Ablation of the Visual Concept Evolution}
\begin{figure}[h!]
\centering
%   \includegraphics[width=8cm]{includes/images/media/w_and_wo_cloning.pdf}%
%   \caption{\label{fig:cloningWandWO} Show learnt parts with and without cloning}
  
  \centering
    % \begin{subfigure}[b]{.49\linewidth}
  \subfloat

    {\includegraphics[width=5.cm]{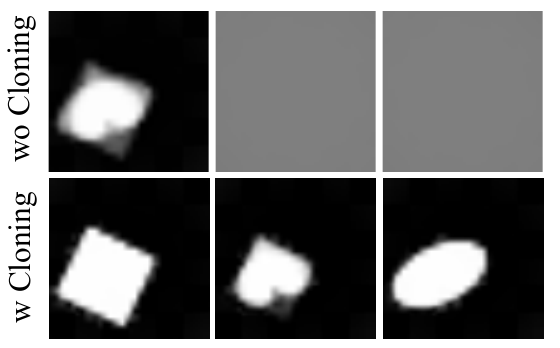}}
        % \includegraphics[width=\linewidth]{images/emoji_interpolation.pdf}
        % \caption{\label{fig:w_and_wo_cloning} Without and With cloning}
    % \hfill
    % \begin{subfigure}[b]{.25\linewidth}
    %     \includegraphics[width=\linewidth]{images/icons_interpolation.pdf}
    % \end{subfigure}
    % \hfill
  \qquad
  \subfloat
    % \begin{subfigure}[b]{.47\linewidth}
     {\includegraphics[width=5.cm]{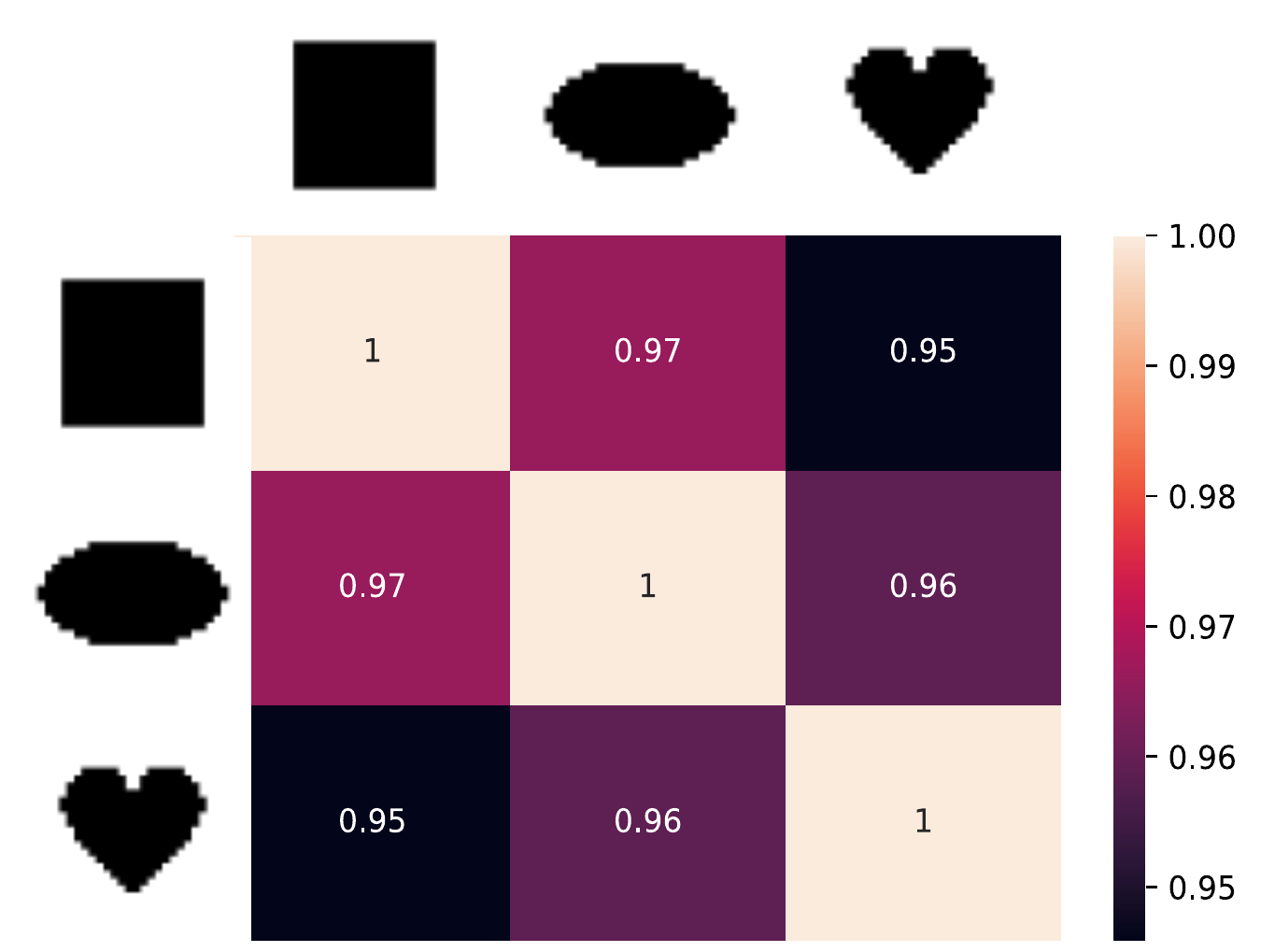}}
        % \caption{\label{fig:confusion_matrix} Confusion Matrix}
    % \end{subfigure}
    %
    \vspace{3pt}

    \caption{\label{fig:w_and_wo_cloning_confusion_m} 
        Effect of cloning. Without cloning~(left-top) our algorithm produces an average visual concept as a proxy to collectively represent similar shapes. However, with cloning~(left-bottom) our algorithm can specialize to pickup subtle differences among the visual concepts, producing individualized concepts. On the right, we show the confusion matrix to show how similar the ground truth visual concepts are, with $1$ denoting perfect similarity.% (right) Confusion Matrix.
    }
\end{figure}

In Fig~\ref{fig:w_and_wo_cloning_confusion_m}, we demonstrate the necessity for the visual concept evolution in our pipeline. Without evolution, a single optimized concept may average multiple similar ground truth concepts. Using evolution, we allow this average concept to split into multiple more specialized child-concepts, that each approximate fewer ground truth concepts. After a few evolution steps, each leaf concept eventually represents a single ground truth concept.
%When fit without cloning, ground truth concepts that are close to each other are assignment to the same concept during fitting. Therefore we might end up learning a average of the ground truth concepts. But when we clone the concepts we give the concepts an ability to specialize resulting in better convergence of the estimated concepts. 

\section{ARI calculation}
The Adjusted Rand Index (ARI) measures the similarity between two clusterings. We use it to compare the decomposition found by our method to the ground truth decomposition. In images without occlusions or with visual concepts that have the same constant color, any layer ordering results in the same reconstructed image, thus the layer order is ambiguous. In cases with ambiguous ordering, we select the layer ordering that gives the highest ARI score.
% In the Multi-dSprites dataset because of the lack of color ques for a given set of elements any layer ordering would lead to the sample final image. We account for this invariance by selecting the layer ordering that gives the highest ARI values. 

\section{Full visual concepts}
\begin{figure}
\centering
  \includegraphics[width=\linewidth]{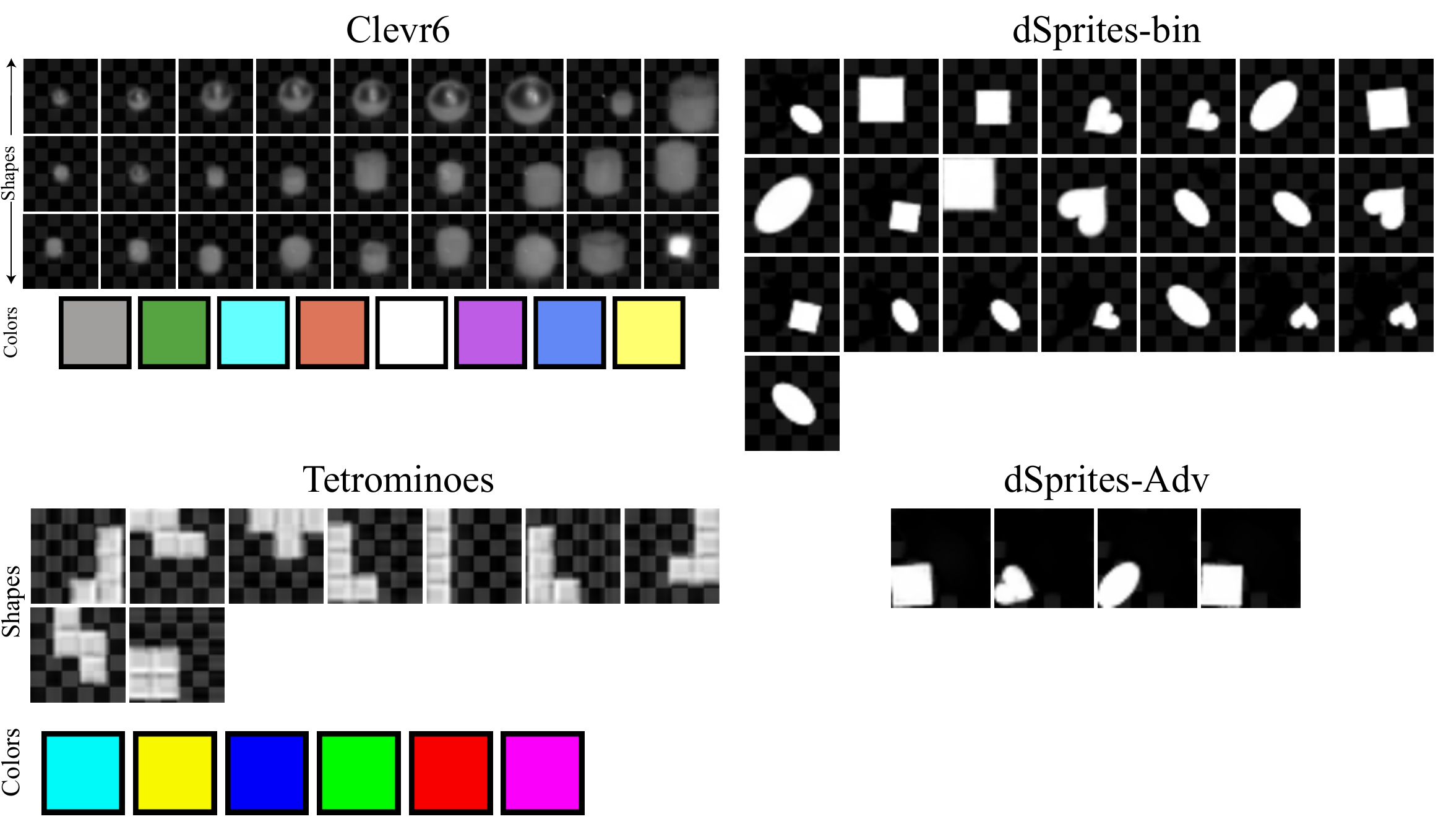}%
  \caption{\label{fig:full_results} All the learned concepts on Tetrominoes, Multi-dSprites bin, Multi-dSprites Adv and Clevr6 datasets.
  }
\end{figure}

\begin{figure}
\centering
  \includegraphics[width=0.92\linewidth]{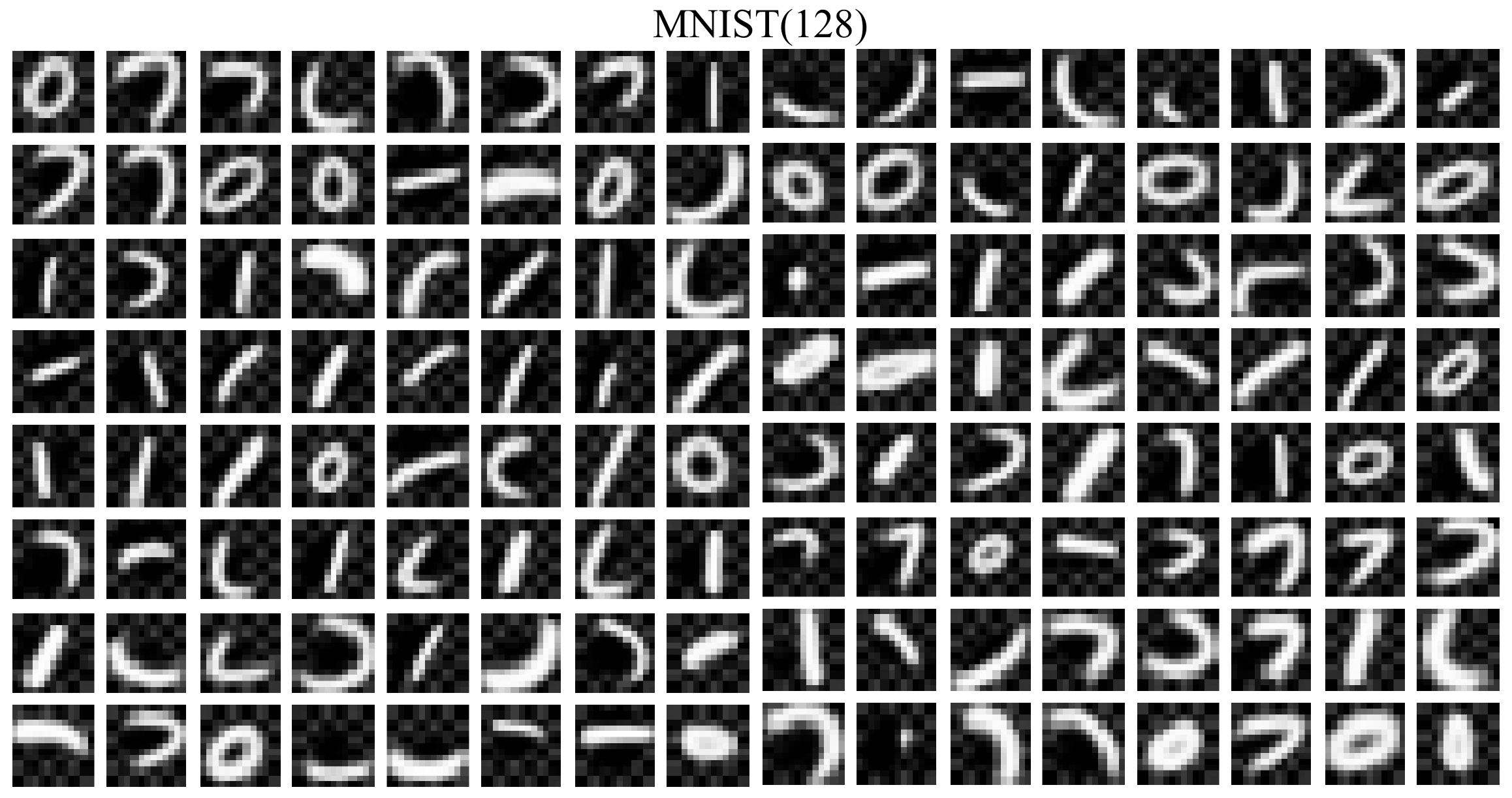}%
  \quad
  \includegraphics[width=0.92\linewidth]{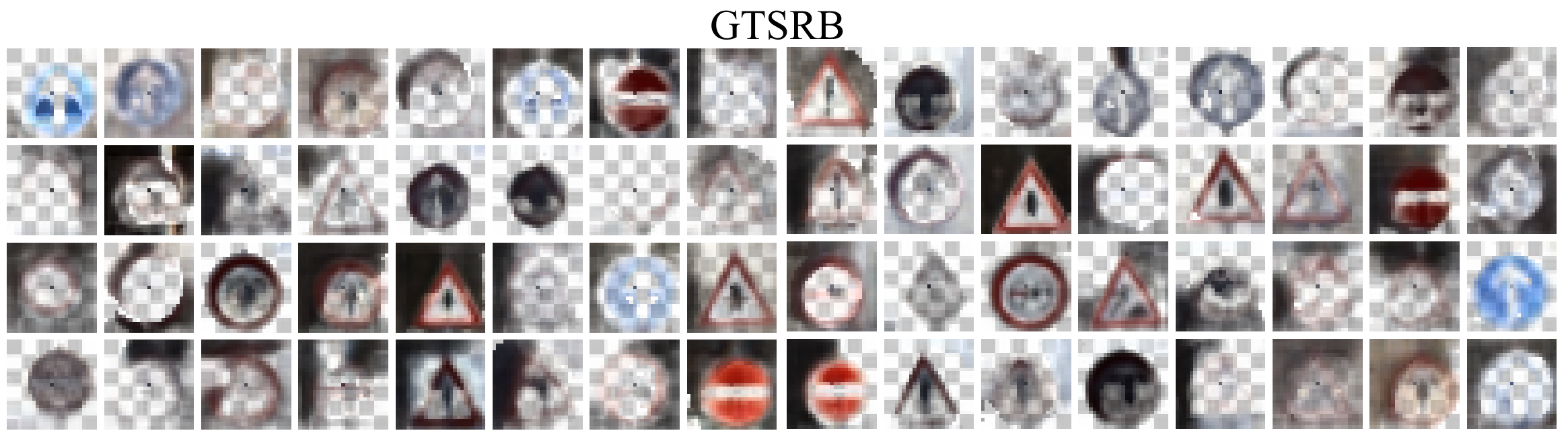}%
  \vspace{3pt}
  \caption{\label{fig:full_results2} All the learned concepts on MNIST and GTSRB datasets.
  }
\end{figure}

In Figures~\ref{fig:full_results} and~\ref{fig:full_results2}, we show all visual concepts learned by our method from each 2D dataset used in the main paper (all visual concepts of the 3D dataset are shown in Figure 10 of the main paper). Additionally, we show visual concepts obtained from the GTSRB traffic sign dataset~\cite{stallkamp2012man} in Figure~\ref{fig:full_results2}, bottom. 

% \begin{figure}
% \centering
%   \includegraphics[width=0.92\linewidth]{includes/images/media/mnist_with_gtsrb.pdf}%
%   \caption{\label{fig:full_results2} All the learned concepts on MNIST and GTSRB datasets.
%   }
% \end{figure}

\section{dSprites Adv. Dataset Details}
\begin{figure}
\centering
  \includegraphics[width=\linewidth]{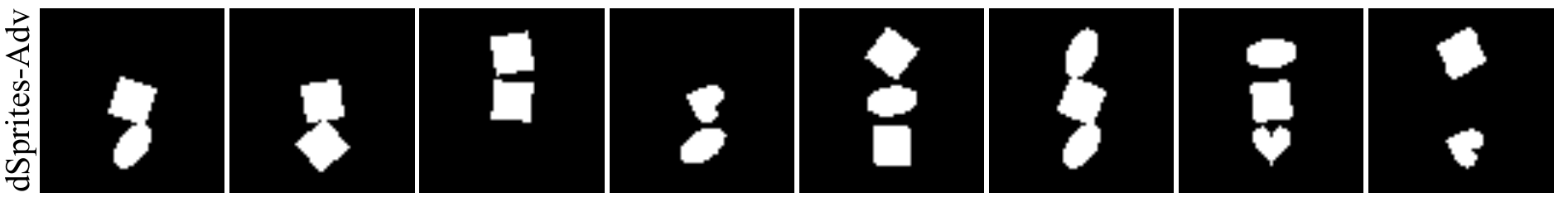}%
  \vspace{3pt}
\caption{\label{fig:adv_sample} Random samples from the dSprites Adversarial Dataset.
  }
\end{figure}

To create the dSprites Adversarial dataset, we place two or three visual concepts in each image. Each concept is placed at one of three pre-defined locations on the canvas (without overlaps). All concepts have the same scale and a random rotation. Figure~\ref{fig:adv_sample} shows samples of the dataset.
%Here we sample three location on the canvas. All the elements used to create the dataset are of the same scale. For each element is randomly rotate between [0,2$\pi$] and place one of the three locations. Each image has atleast two elements. 

\section{Background Handling}
When using alpha-compositing, we treat the background as a special layer $L_n$ that is locked to the back of the layer stack (i.e. it is occluded by all other layers when using alpha-compositing) and does not have layer parameters. The background is represented by a special visual concept $V_m$ that is only used by the background layer and is initialized with a constant value of $0.5$ in all pixels. During optimization of a given image $\tilde{I}$, we optimize the background visual concept before the other concepts or layers, to make sure the other layers don't represent parts of the background.
%the last layer in compositing process. During the updating the layer parameter phase we first search for the background and then the layers that are on top of the background. A background layer is initialize with a constant value of 0.5 and then updated during the optimization process similar to the visual concepts.   

\section{3D Scene Reconstruction segmentation}
We also measure the quality of our decompositions by comparing the 2D projections of the segmented 3D scene to a known ground truth. We achieve an ARI of 99.3\% on this task.

\section{Video}
In the supplementary material, we include a video that visualizes the optimization process of our method, showing the optimized visual concepts in each iteration and the layer segmentation of the reconstructed image. For clarity, we demonstrate the optimization on a dataset consisting of a single image with multiple repeating visual concepts.
%on a pattern image. The video shows the progression of the concepts with every iteration. The segmentation reconstruction show which concept is used at a given local. The segmentation reconstruction should also give the viewer a sense for the estimated occlusion ordering. 

\section{Code}
We include a development version of our code in the supplemental that can be used to reproduce the results.

\end{document}